\documentclass[twoside,11pt]{article}

\usepackage[preprint,abbrvbib]{jmlr2e}
\usepackage[utf8]{inputenc}
\usepackage{tikz-cd}
\jmlrheading{1}{2020}{1-38}{4/00}{10/00}{cantelobre21}{Théophile Cantelobre, Benjamin Guedj, María Pérez-Ortiz, John Shawe-Taylor}
\ShortHeadings{PAC-Bayesian Structured Prediction with ILE}{Cantelobre, Guedj, Pérez-Ortiz, Shawe-Taylor}
\firstpageno{1}
\usepackage{color}
\usepackage{diagbox}
\usepackage{latexsym}
\usepackage{amsmath}
\usepackage{amsfonts}
\usepackage[mathcal]{eucal}
\usepackage{multirow}
\usepackage{xspace}
\usepackage[ruled,vlined,linesnumbered]{algorithm2e}
\usepackage{adjustbox}
\newcommand{\github}{\href{https://github.com/theophilec/PAC-Bayes-ILE-Structured-Prediction}{https://github.com/theophilec/PAC-Bayes-ILE-Structured-Prediction}\xspace}

\newcommand{\norm}[1]{\left\Vert #1 \right\Vert}

\newcommand{\expect}[1]{\mathop{\mathbb{E}}_{#1}}

\newcommand{\grad}{\nabla}
\DeclareMathOperator{\Tr}{Tr}
\DeclareMathOperator{\Cov}{Cov}

\makeatother

\usepackage{amsthm}
\usepackage{cleveref}
\newcommand{\BlackBox}{\rule{1.5ex}{1.5ex}}  
\renewenvironment{proof}{\par\noindent{\bf Proof\ }}{\hfill\BlackBox\\[2mm]}
\newtheorem{example}{Example} 
\newtheorem{theorem}{Theorem}

\newtheorem{remark}[theorem]{Remark}

\newtheorem{definition}[theorem]{Definition}

\newtheorem{lemme}{Lemma}
\newtheorem{prop}[theorem]{Proposition}

\newtheorem{problem}{Problem}

\crefname{problem}{Problem}{Problems}
\crefname{theorem}{Theorem}{Theorems}
\crefname{prop}{Proposition}{Propositions}
\crefname{equation}{Eq.}{Eqs.}
\crefname{figure}{Figure}{Figures}
\crefname{section}{Section}{Sections}
\crefname{table}{Table}{Tables}
\crefname{lemme}{Lemma}{Lemmas}
\crefname{corollary}{Corollary}{Corollaries}
\crefname{example}{Example}{Examples}
\crefname{appendix}{Appendix}{Appendixes}
\crefname{remark}{Remark}{Remark}

\begin{document}

\title{A PAC-Bayesian Perspective on Structured Prediction with Implicit Loss Embeddings} 
\author{%
\name Th\'eophile Cantelobre \\
\addr Mines ParisTech -- PSL, Inria and Sorbonne Université\\ France
\AND
\name Benjamin Guedj \\
\addr Inria and University College London \\ France and United Kingdom
\AND
\name María Pérez-Ortiz\\
\addr University College London \\ United Kingdom
\AND
\name John Shawe-Taylor\\
\addr University College London \\ United Kingdom
}

\editor{}
\maketitle

\begin{abstract}%
Many practical machine learning tasks can be framed as Structured prediction problems, where several output variables are predicted and considered interdependent. Recent theoretical advances in structured prediction have focused on obtaining fast rates convergence guarantees, especially in the Implicit Loss Embedding (ILE) framework. PAC-Bayes has gained interest recently for its capacity of producing tight risk bounds for predictor distributions. This work
proposes a novel PAC-Bayes perspective on the ILE Structured prediction framework. We present two generalization bounds, on the risk and excess risk, which yield insights into the behavior of ILE predictors. Two learning algorithms are derived from these bounds. The algorithms are implemented and their behavior analyzed, with source code available at \href{https://github.com/theophilec/PAC-Bayes-ILE-Structured-Prediction}{https://github.com/theophilec/PAC-Bayes-ILE-Structured-Prediction}.
\end{abstract}


\section{Introduction}

Structured prediction (also referred to as Structured output prediction) is a widely studied problem in machine learning of great theoretical and practical interest. Broadly, we define Structured prediction as the joint prediction of interdependent or constrained decision variables. 

Although we make this definition precise below, let us first present an example of a recent application of Structured prediction: Scene Graph Generation. Given an image, a goal can be to extract the semantic information that describes the depicted scene, that is, to detect objects present in the image and the semantic relationships between them. Formally, the task consists in predicting a graph of all entities and the relations between then from an image (for example, \emph{the man is wearing a hat}). Note that this graph is on the instance level, so \emph{the man wearing the hat}, is known to be \emph{the same one holding a bucket} (which \emph{the horse is known to be eating from}).
 
The decision variable for this task is the scene graph. Seen as a discrete object, its domain is combinatorially large. Although it would be possible to first predict instances independently, and then the relations between pairs of detected instances, this strategy clearly appears sub-optimal. 

Imagine for example that the entities predicted are, for whatever reason, a dog and a lion. Jointly predicting the entities avoids predicting particularly improbable pairs. On the other hand, even if not all entity and relation triplets are represented in the training data, the correlations between objects should hopefully be taken into account. For example, even if the particular configuration of a horse eating from a bucket is not in the training data, all other domestic animal feeding situations are correlated with it (man feeding cow with a bucket,
man feeding a horse with another object, etc.). We refer the interested reader to \cite{surrey} which provides a complete survey of the field of Visual Semantic Information Pursuit.

Beyond practical applications, prediction of joint or constrained variables is of great theoretical interest and encompasses tasks such as ordinal regression, multiclass classification, multilabel classification, or manifold regression, to cite but a few. In fact, we will see that most machine learning problems can be formulated as a form of Structured output prediction problem, from binary classification to ordinal regression. Furthermore, many theoretical and practical challenges of
statistical learning are present in Structured output prediction:
\begin{enumerate}
    \item Both training and inference are NP-Hard in general, making large-scale (in problem size, and dataset size) problems hard to solve. Indeed, many common approaches to optimization rely on repeatedly performing inference on (part of) the training set \citep[for example, the Structured SVM algorithm, see][]{ssvm}.
    \item Seen as a classification problem with a very large label space, Structured output prediction has few statistical or algorithmic guarantees with respect to standard approaches. In fact, as highlighted by \cite{nowak-vila-general}, many algorithms are known to be inconsistent.
\end{enumerate}

In this work, we present a PAC-Bayes analysis of the Implicit Loss Embedding (ILE) approach to Structured prediction. In particular we concentrate of studying the generalization and consistency properties of the ILE approach.

\paragraph{Generalization.}
The former is a central problem in statistical learning theory and quantifies how well we can expect the performance of a predictor trained on a finite dataset to generalize to the entire data-generating distribution. There would be too many seminal contributions to the study of generalization to cite -- we refer to \cite{devroye1997} and \cite{Vapnik1998} and references therein.

In this work, we consider the PAC-Bayes learning paradigm to study generalization. A generalization of the Probably Approximately Correct framework of \cite{valiant-pac}, PAC-Bayes bounds originate in the seminal papers of \cite{STW1997,McAllester1998,McAllester1999} and further developed by \cite{seeger2002,catoni2004statistical,catoni2007}. In this framework, instead of considering the generalization qualities of a given predictor, we study that of a distribution of predictors, learned from the training data. The bounds we aim to obtain make little to no hypotheses on the data-generating distribution. There has been a surge of interest in PAC-Bayes in the past few years, as it has re-emerged as a promising framework for assessing generalization performance in several settings, such as coherent risk measures \citep{mhammedi2020pacbayesian}, deep neural networks \citep{DBLP:conf/uai/DziugaiteR17,letarte19dichotomize}, differential privacy \citep{DBLP:conf/nips/Dziugaite018,DBLP:conf/icml/Dziugaite018}, meta-learning \citep{DBLP:conf/icml/AmitM18}
or contrastive learning \citep{nozawa2019pacbayesian} to name but a few.
We refer to \cite{guedj_primer} for a recent survey and to \cite{GueSTICML} for a recent tutorial on PAC-Bayes.
We outline some contributions of PAC-Bayesian theory to Structured prediction in the \emph{Prior work} paragraph below.

\paragraph{Consistency.}
In statistical learning theory, much effort has been spent solving the intractability of minimizing the empirical $0-1$ loss. We will make the definition of consistency precise in what follows. Broadly, a surrogate method is consistent for a given task (for example, minimizing the empirical $0-1$ loss) if minimizing the surrogate loss (for example, the quadratic loss) solves the task. For general treatments about consistency, see for example \cite{devroye1997} and \cite{bartlett}.

In the rest of the paper, we introduce and reason around Fisher consistency and its implications in practice.

\paragraph{Prior work.}
There have been many attempts to tame the Structured output prediction problem, both from a statistical and an algorithmic point of view. These include for example many \emph{ad hoc} approaches such as Structural Support Vector Machines \citep{ssvm} or Conditional Random Fields \citep{lafferty_crf,taskar_mmn}. \cite{sp-cv} provides a survey of applications of Structured prediction for Computer vision. \cite{wainwright_graphical} presents a unified approach to learning and inference over graphical models. Many ensemble methods based on building efficient subgraphs over graphical models have been developed: for example, \cite{simplemkl}, \cite{bach_mkl}, \cite{argyriou_convex_2008}, \cite{marchand_spanning}. 

Several quadratic methods have focused on (consistently) training quadratic surrogate methods for the Structured learning problem. These include Kernel Dependency Estimation \citep[KDE, see][]{weston_kde,cortes_kde}, Input-Output Kernel \citep[for example,][]{brouard} and ILE approaches \citep{ciliberto2016,osokin,rudi_manifold,nowak-sharp-analysis,localized-sp}. Non-quadratic methods also exist \citep{nowak2020consistent}. For instance, \cite{nowak-vila-general} generalizes the ILE framework to a more general class of surrogates.

Several authors have proposed PAC-Bayes approaches to Structured prediction. Let us cite just three works which show the diversity of the PAC-Bayesian approach: \cite{marchand_spanning} uses PAC-Bayes arguments to justify an approximation strategy for graphical models, while \cite{mcallester_structured_2} provides generalization bounds with a focus on tasks involving language models. Finally, \cite{giguere_2013} provides a PAC-Bayesian generalization bound for the KDE algorithm, which we will return to.
For completeness we also mention recent works focusing on the decoding problem (that is, constructing a solution to the task from the surrogate solution), even though this falls outside of the scope of this paper \citep{mensch_blondel,blondel_oracles}.

\paragraph{Contributions.}
In this work, we propose an original PAC-Bayes analysis of the ILE framework in~\cite{ciliberto_general_2020}. For most of the paper, we concentrate on the multi-label problem. In particular:
\begin{itemize}
    \item We provide a PAC-Bayes generalization bound using classic PAC-Bayes arguments based on~\cite{pb_zhang}.
    \item We extend the Comparison inequality in \cite{ciliberto_general_2020} to propose a new bound on the expected excess risk that depends on the performance of the predictor distribution on the training set. This result holds with high probability over the training set, thanks to recent work on PAC-Bayes generalization bounds in \cite{haddouche2020} for unbounded loss functions, such as commonly used regression losses. 
    \item We thoroughly analyze the bound and its behavior, allowing us to illustrate the importance of using a structured loss (for example, the Hamming loss) instead of the $0-1$ loss. This yields theoretical insights as well as practical advice for Structured prediction.
    \item Finally, we present learning algorithms derived from our novel PAC-Bayes bound and test their performance on Structured output prediction datasets. All our experiments are reproducible from the associated GitHub repository\footnote{\github}.
\end{itemize}

The rest of the paper is organized as follows: in \cref{sec:ile}, we introduce the Structured learning formulation and the ILE framework. In \cref{sec:pbclass}, we present our PAC-Bayes classification bound for ILE Structured output prediction. In \cref{sec:consistency}, we derive and analyze a novel PAC-Bayes generalization bound providing insights into the consistency of ILE methodology. In \cref{sec:algorithms}, we solve the multi-label classification problem on Structured output
prediction datasets using the derived result, through two methods. In \cref{sec:discussion} we put our results into perspective with respect to the ILE framework. Finally, we gather our experimental results in \cref{sec:experiments}.

\section{Structured output Learning with Implicit Loss Embeddings }\label{sec:ile}

We now introduce our notation and the Implicit Loss Embedding (ILE) framework for Structured output prediction.

\subsection{Supervised Learning \& Excess risk }\label{sec:supervised}
Let $\mathcal X$ be an input space equipped with a kernel $k(x, x^\prime) = \langle \phi(x), \phi(x^\prime)\rangle$ for any $x,x^\prime\in\mathcal{X}$. Let $\mathcal Y$ be the label space for the training data, and $\mathcal Z$ denotes the output space. In many cases, $\mathcal Y = \mathcal Z$, but the distinction is necessary, for example, in information retrieval. Indeed in this setting, $\mathcal X$ can be for example the space of search engine queries with $\mathcal Z$ an ordering over documents presented in response to a query and $\mathcal Y$ a relevance score for each result for the query (instead of a ranking). Let $\Delta: \mathcal Z \times \mathcal Y \rightarrow \mathbb R$ be a non-negative loss function. 

We aim to solve the supervised learning problem, \emph{i.e.}, given a dataset $\left\{(x_i, y_i)\right\}_{i=1}^m$ drawn i.i.d. from a data-generating distribution $\rho$ over $\mathcal X \times \mathcal Y$, find a measurable function $f_m: \mathcal X \rightarrow \mathcal Z$ that minimizes the expected task risk (or Bayes risk) defined by:
\begin{equation}
\mathcal E (f) = \int_{\mathcal X \times \mathcal Y}{\Delta(f_m(x), y)\mathrm{d}\rho(x, y)}\label{eq:expectedtaskrisk}.
\end{equation}
We define $f^*$ the optimal predictor given by:
\begin{equation}
f^* = \arg\min_{f_m:\mathcal X \rightarrow \mathcal Z}\mathcal E (f_m),
\end{equation}
where $\mathcal X \rightarrow \mathcal Z$ denotes all measurable functions from $\mathcal X$ to $\mathcal Z$.

We define the \emph{expected excess risk} $\mathcal E(f_m) - \mathcal E(f^*)$ between $f_m$ and $f^*$, which characterizes the sub-optimality of a predictor $f_m$ with respect to the task loss $\Delta$ and data-generating distribution $\rho$.

In what follows, we present the ILE framework of~\cite{ciliberto_general_2020} as well as some selected results. We refer to~\cite{ciliberto_general_2020} and references therein for a more complete treatise.

\subsection{Implicit Loss Embedding }

\begin{figure}[t]
    \centering


    \adjustbox{width=0.6\textwidth}{
\begin{tikzcd}
\mathcal F \arrow[rr, "h"] &  & \mathcal H \arrow[dd, "d"', bend right] & \\
&  & &  \\
\mathcal X \arrow[rr, "f=d\circ g", bend right] \arrow[uu, "X"] \arrow[rruu, "g"] &  & \mathcal Z \arrow[uu, "\psi" description, bend right] \arrow[r, "{\Delta=\langle\psi, \varphi\rangle}"', no head, dotted, bend right] & \mathcal Y \arrow[luu, "\varphi"', shift right]
\end{tikzcd}}
\caption{Summary illustration of the different sets and maps in the ILE framework.}
\label{fig:ile-schematic}
\end{figure}
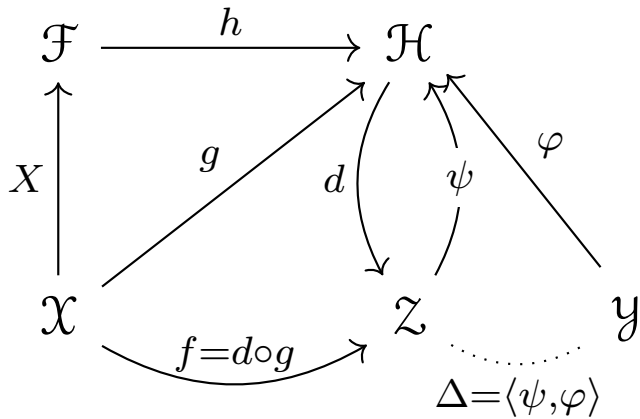

The central idea to the ILE framework is the existence of an embedding of the output and label spaces such that the loss function $\Delta$ is the dot product of the embedding functions to this space (similarly to the idea behind the use of kernels). Formally, we define an Implicit Loss Embedding for $\Delta$ as follows. See \cref{fig:ile-schematic} for a summary illustration of the different involved sets and maps.

\begin{definition}
[Implicit Embedding, ILE]
A loss function $\Delta: \mathcal Z \times \mathcal Y \rightarrow \mathbb R$ has an Implicit Loss Embedding (ILE) if there exists a Hilbert space $\mathcal H$ and two measurable maps $\varphi: \mathcal Y \rightarrow \mathcal H$ and $\psi: \mathcal Z \rightarrow \mathcal H$ such that
\begin{equation}
\forall (z, y)\in\mathcal Z \times \mathcal Y,~\Delta(z, y)=\langle \psi(z), \varphi(y)\rangle.
\end{equation}
\end{definition}

Let us motivate the introduction of this definition. First, notice that most structured (and indeed unstructured) loss functions fall into this formalism. For example, if $\mathcal Y$ and $\mathcal Z$ are finite, then the ILE is the matrix formed by the values of the loss function for $(y,z)$ pairs, encoded as one-hot encodings $e_y$ and $e_z$. In \cref{ex:example}, we show that the Hamming loss for multi-label classification has an ILE. We offer a formal definition of the
$\ell$-multilabel binary classification problem.
\begin{problem}
    [$\ell$-multilabel binary classification problem]
    \label{problem:multilabel}
    With $\ell\in\mathbb N^*$, let $\mathcal Y = \lbrace 0, 1\rbrace^\ell$. Within the formalism described in \cref{sec:supervised}, the problem is to learn a predictor $f: \mathcal X \rightarrow \mathcal Y$ from training data sampled iid from the data-generating distribution $\rho$ over $\mathcal X \times \mathcal Y$.
\end{problem}

Some generic examples of $\ell$-multilabel binary classification problems include for example detecting several characteristics of input datum $x$ (i.e. different objects present in an image, topics of a text, presence of different defects). Using a similar example as in the introduction, detecting several objects in an image \emph{should} be easier if the objects are jointly predicted.

\begin{example}\label{ex:example}
    In binary $\ell$-multi-label learning, the Hamming loss $H$ is ILE where
\begin{equation}\label{eq:hamming_def}
    H(z, y) = \frac{1}{\ell}\sum_{k=1}^\ell\mathbb I(z_k \neq y_k) = 1 - \frac{1}{\ell}\sum_{k=1}^\ell \mathbb I(z_k = 0) \mathbb I(y_k = 0) + \mathbb I(z_k = 1)\mathbb I(y_k=1).
\end{equation}
Furthermore, $\vert \mathcal Y \vert = 2^\ell$ and $\dim(\mathcal H) = 2\ell + 1$.
\end{example}

Different authors have shown that a wide variety of machine learning problems fall into the ILE framework. As a matter of fact,~\citet[Sec. 6]{ciliberto_general_2020} provide a collection of simple sufficient conditions for a learning problem to admit an ILE (the finite cardinality of $\mathcal Y$ and $\mathcal Z$ is the simplest of such conditions).

Second, the ``separability'' of $\Delta$ as illustrated in \cref{eq:f-star} ensures that the learning problem can be (Fisher-)consistently solved by solving a more computationally efficient quadratic regression problem, approximating $\varphi(y)$ from $x$. In particular, this ensures that, in structured contexts, the so-called pre-image problem is not solved during training as it is in Structured SVM methods.

In what follows, we assume that $\Delta$ is ILE with associated maps $\psi$ and $\varphi$. We also assume that $\mathcal Y$ and $\mathcal Z$ are finite sets, and that $\mathcal H$ is finite dimensional. In specific examples, we will return to the $\ell$-multi-label binary classification problem with the Hamming loss, as defined in \cref{problem:multilabel} and \cref{eq:hamming_def}.

\subsection{Quadratic surrogate and consistency }\label{sec:defconsistency}
Under the assumptions that the task loss is ILE and that the data-generating distribution $\rho$ can be factorized as $\rho(x, y) = \rho(x) \rho(y|x)$,~\cite{ciliberto_general_2020} prove the following results, which naturally lead to a consistent surrogate method (in a sense we define below). We briefly survey this approach to set the context and demonstrate the utility of such a framework, in theory and in practice.

\cite{ciliberto_general_2020} first prove that the pointwise conditionally optimal predictor is identically equal to the minimizer of \cref{eq:expectedtaskrisk}:
\begin{lemme}
[\citealp{ciliberto_general_2020}]
Assume that the data-generating distribution $\rho$ can be factorized as $\rho(x,y) = \rho(y|x)\rho_\mathcal X(x)$, then
\begin{equation}
f^*(x) = \arg\min_{z\in\mathcal Z}\int_\mathcal Y \Delta(z, y)d\rho(y|x).
\end{equation}
\end{lemme}

From this, because $\Delta(z, y) = \langle\psi(z), \varphi(y)\rangle$ is linear in $\varphi(y)$, the optimal predictor $f^*$ is naturally written as a function of the conditional mean embedding of $\varphi(y)$ given $x$, which we denote $g^*(x) = \int_\mathcal Y \varphi(y)d\rho(y|x)$, \emph{i.e.},
\begin{equation}\label{eq:f-star}
f^*(x) = \arg\min_{z\in\mathcal Z}\langle\psi(z), g^*(x)\rangle.
\end{equation}
Thus, in order to estimate $f^*$, it is natural to seek to estimate $g^*$ by $g$ and ``plug-in'' $g$ to define $f$ as 
\begin{equation}
\forall x\in\mathcal X, f(x) = \arg\min_{z\in\mathcal Z}\langle\psi(z), g(x)\rangle.\label{eq:predictordef}
\end{equation}
It can be shown, see for example \cite{gretton_cme}, that $g^*$ is the minimizer of the following expected quadratic risk:
\begin{equation}\label{eq:Rg}
\mathcal R(g) = \int_{\mathcal X \times \mathcal Y}\norm{\varphi(y) - g(x)}^2d\rho(x,y).
\end{equation}
A consistent empirical counterpart of \cref{eq:Rg} for a given sample is:
\begin{equation}\label{eq:R-n-g}
\mathcal R_m(g) = \frac{1}{m}\sum_{i=1}^m\norm{g(x_i) - \varphi(y_i)}^2 + \lambda \norm{g}^2,
\end{equation}
where $\lambda$ is a regularization parameter. There exist $\alpha_1, \ldots, \alpha_m$ $n$ functions in $\mathcal X \rightarrow \mathcal H$ such that $g_m$ minimizes $\mathcal R_m$ where $g_m$ is defined by 
\begin{equation}
g_m(x) = \sum_{i=1}^m \alpha_i(x)\varphi(y_i).
\end{equation}
The associated classifier is thus defined by:
\begin{align}
\forall x\in\mathcal X, ~f_m(x) &= \arg\min_{z\in\mathcal Z}\langle \psi(z), g_m(x)\rangle \\ 
                                &= \arg\min_{z\in\mathcal Z}\sum_{i=1}^m \alpha_i(x)\langle \psi(z), \varphi(y_i)\rangle.
\end{align}
The above computation is called the ``loss trick'' by \cite{ciliberto_general_2020} in analogy to the well-known ``kernel trick''. This proves that the existence of the ILE embedding is enough and that knowledge of $\varphi$ and $\psi$ are not required.

\paragraph{Efficiently solving the decoding problem.}
As we have outlined above, the general structure of the ILE learning method is to solve a quadratic surrogate regression problem, and then at inference solve a decoding problem (the $\arg\min$ in \cref{eq:f-star}). A priori, this can be solved in $O(\vert\mathcal Z\vert)$ time. Because $\vert \mathcal Z\vert$ is combinatorial in nature (for example $\vert \mathcal Z \vert = 2^\ell$ for the $\ell$-label binary classification problem) this can be costly and even impossible at large
scale. 

In this work, we focus on the link between the plug-in classifier $f_m$ and the surrogate regressor $g_m$ without worrying about how the $\arg\min$ is computed. The problem of efficiently solving the decoding problem is the subject of some recent work in, for example, \cite{blondel_oracles, mensch_blondel}.

\paragraph{Fisher consistency}\cite{ciliberto_general_2020} show that estimating $g$ is \emph{Fisher consistent}, \emph{i.e.}, that if the optimal regressor $g^*$ is found, then the associated predictor $f^*$ is optimal for $\mathcal E$. Formally, there exists a map $d: \mathcal H \rightarrow \mathcal Z$ (for example, the $\arg\min$ function over $\mathcal Z$) such that $\mathcal E(f^*) = \mathcal E(d\circ g^*)$. 

They prove a stronger result, upper-bounding the expected excess prediction risk $\mathcal E(f) - \mathcal E(f^*)$ as a function of the expected excess regression risk $\mathcal R(g) - \mathcal R(g^*)$, their \emph{comparison inequality}. This result will be the basis for our own analysis in \cref{sec:consistency}. For easy cross-referencing, we present a stronger version of the inequality obtained by withholding the last step in the proof of the inequality in~\cite{ciliberto_general_2020}.

\begin{theorem}
[Strong Comparison Inequality]\label{thm:strongcomp}
Assume $\mathcal Z$ is compact and $\Delta$ has an ILE. Let $g: \mathcal X \rightarrow \mathcal H$  be \textbf{measurable}, and $f$ defined as in \cref{eq:predictordef}. Then, 
\begin{equation*}
    \mathcal E(f) - \mathcal E(f^*) \leq 2c_\Delta\int_\mathcal X \norm{g(x)-g^*(x)}_\mathcal H d\rho_{\mathcal X} (x),
\end{equation*}
where $c_\Delta = \sup_{z\in\mathcal Z}\norm{\psi(z)}$ and $\rho_\mathcal X$ is the data-generating distribution marginalized over $y$.

\end{theorem}
In particular, this implies the Comparison inequality (\cref{thm:comp}).

\begin{theorem}
    [Comparison inequality, Thm. 3, \citealp{ciliberto_general_2020}]
\label{thm:comp}
Assume $\mathcal Z$ is compact and $\Delta$ has an ILE. Let $g: \mathcal X \rightarrow \mathcal H$ be \textbf{measurable}, and $f$ defined as in \cref{eq:predictordef}. Then, 
\begin{equation*}
\mathcal E(f) - \mathcal E(f^*) \leq 2c_\Delta \sqrt{\mathcal R(g) - \mathcal R(g^*)},
\end{equation*}
where $c_\Delta = \sup_{z\in\mathcal Z}\norm{\psi(z)}$.
\end{theorem}

Note that such calibration functions exist for a wider range of surrogates, see \cite{nowak-vila-general} for a more general approach. These results are analogous to the $\psi$-transform bounds in, for example, \cite{bartlett}.

In this section, we presented the supervised Structured learning problem and the ILE framework. We also introduced the relevant results for the rest of the paper. We now present the PAC-Bayes approach, as well as prove our first results.

\section{A PAC-Bayes take on ILE classification}\label{sec:pbclass}

In this section, we offer a novel perspective on the learning problem described in \cref{sec:ile} using tools from the PAC-Bayes framework. We first outline the main tenets of the PAC-Bayes approach, and a classic PAC-Bayes bound. Using these tools we prove a generalization bound for ILE classification.

\paragraph{Notation.} In the rest of the paper, we define the empirical task risk as:
\begin{equation}
\mathcal E_m(f) = \frac{1}{m}\sum{}_{i=1}^m \Delta(f(x_i), y_i).
\end{equation}
The expectation of the measurable function $f$ of a random variable $X$ with domain $\mathcal X$ under distribution $\mathcal Q$ is defined as (when it exists):
\begin{equation*}
\expect{X\sim\mathcal Q}f(x) = \int_\mathcal X f(x)d\mathcal Q(x).
\end{equation*}
The Kullback-Leibler divergence between distributions $\mathcal Q$ and $\mathcal P$ is denoted $\mathcal K(\mathcal Q, \mathcal P)$ and defined as
\begin{equation*}
\mathcal K(\mathcal Q, \mathcal P) = \int\log\frac{d\mathcal Q(x)}{d\mathcal P(x)}d\mathcal Q(x).
\end{equation*}
When there is no ambiguity, we identify linear maps between finite-dimensional vector spaces, associated matrices in the canonical bases, and the vectors of matrix coordinates.

Finally, in the rest of the paper, we will make the slight notation abuse of writing $f\sim\mathcal Q$ when $f$ is the structured predictor associated to stochastic regressor $g\sim \mathcal Q$.

\subsection{Canonical PAC-Bayes bounds}
We introduce the PAC-Bayes setting and present a canonical PAC-Bayes generalization bound from \cite{pb_zhang}.
In PAC-Bayes theory, one considers stochastic predictors sampled from a distribution $\mathcal Q$ over the space of measurable maps $\lbrace{ f:\mathcal X \rightarrow \mathcal Z \rbrace}$. Intuitively, this can be understood as sampling a different predictor from a distribution $\mathcal Q$ for each inference task, for example for each data-point in a dataset.

Thus, if learning is generally the process of estimating an optimal predictor, in the PAC-Bayes approach, learning consists in estimating an optimal distribution of predictors. Similarly to the general learning setting, the definition of ``optimality'' depends on the performance of the predictor (distribution) on the training set and on regularization terms. 

Because the optimized distribution $\mathcal Q$ depends on the training set used for ``learning'': in analogy to Bayesian statistics, this distribution is called the posterior. Conversely, we consider a data-independent prior distribution over predictors $\mathcal P$. 

The result below bounds the expected risk over a posterior distribution $\mathcal Q$  of predictors from the empirical risk over the dataset, with high probability over the training set. Note that the bound holds for any data-generating distribution $\rho$. The generalization penalty terms contain the Kullback-Leibler divergence between $\mathcal Q$ and $\mathcal P$. Because the Exponential information inequality presented in \cite{pb_zhang} uses an alternative so-called one-sample formulation, we refer to the corresponding result in \cite{giguere_2013}:
\begin{theorem}
[\citealp{giguere_2013}, Thm. 10.3.]
\label{thm:zhang}
Let $\zeta$ be an arbitrary (in particular, not necessarily bounded) loss function. 

For any data-independent distribution $\mathcal P$, $\delta >0$, and $a > 0$,  with probability at least $1-\delta$ over the training set drawn from $\rho^m$, for any posterior distribution $\mathcal Q$ such that $\mathcal Q \ll \mathcal P$ and $\mathcal P \ll \mathcal Q$,
\begin{equation*}
    -\expect{g\sim Q}\log\expect{x,y \sim \rho}e^{-\zeta(g(x), y)} \leq \frac{1}{m}\left(
\expect{g\sim Q}\sum_{i=1}^m \zeta(g(x_i), y_i)+ \mathcal K(Q, \mathcal P) + \log\frac{1}{\delta}
\right).
\end{equation*}
\end{theorem}
We use \cref{thm:zhang} to prove a first result below.

\subsection{ILE generalization bound}
\begin{figure}[t]
\begin{center}
    \includegraphics[width=0.8\textwidth]{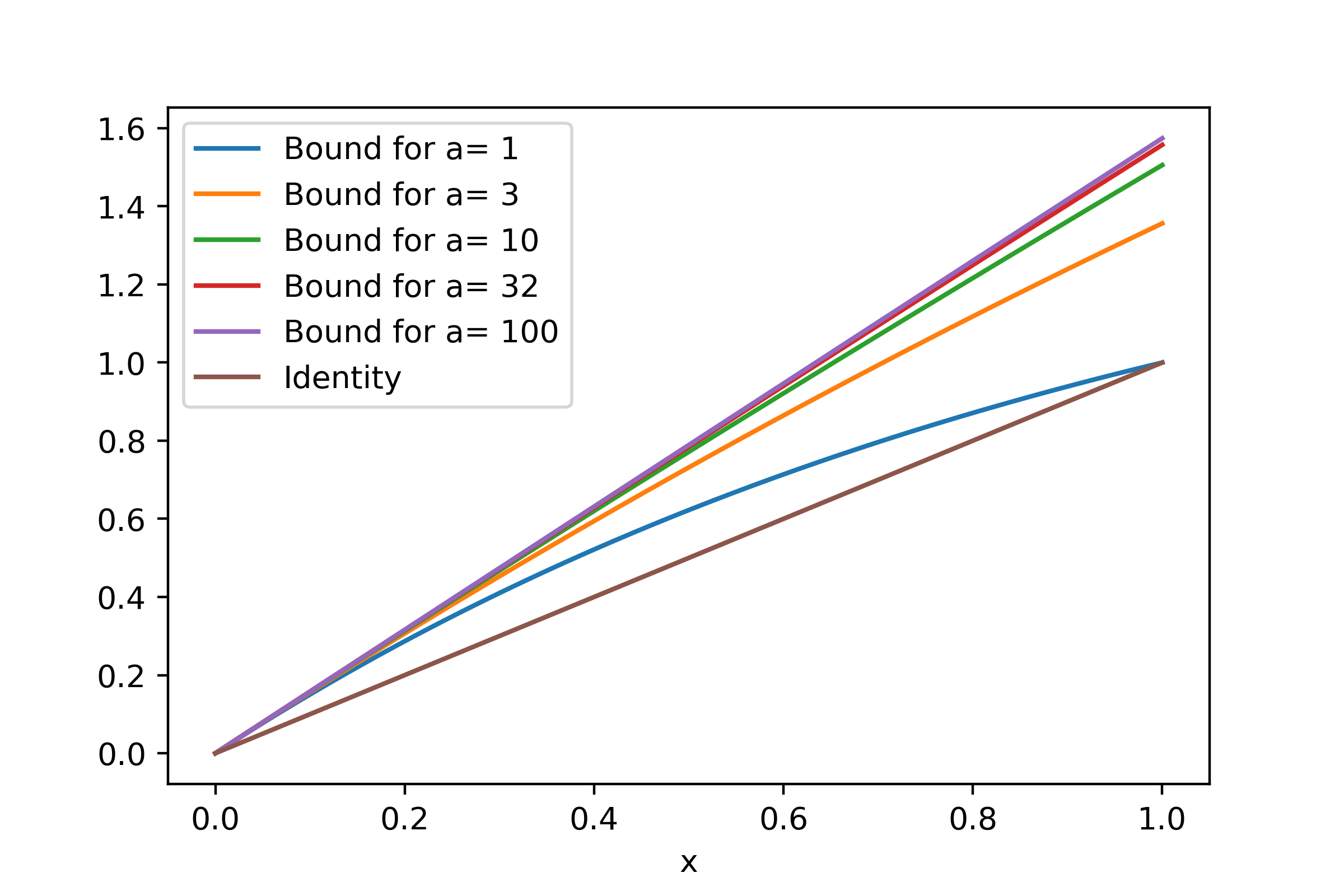}
\end{center}
\caption{Illustration of \cref{prop:idexptemp} (the exponential bound on the identity) for different values of $a>1$.}
\label{fig:id-exp-bound}
\end{figure}

In this section, we prove the following novel result, which applies \cref{thm:zhang} to ILE Structured prediction:
\begin{theorem}
[Classification bound]
\label{thm:ilezhang}
Assume that $\Delta$ admits an ILE defined by $\Delta(z, y) = \langle\psi(z), \varphi(y)\rangle$.
Let $g: \mathcal X \rightarrow \mathcal H$ a measurable function, and $f(x) = \arg\min_{z \in Z} \langle\psi(z), g(x)\rangle$. 

For any data-independent distribution $\mathcal P$, for any $\delta > 0$, for any $\lambda >0$, with probability at least $1 - \delta$ over the training set sampled iid from $\rho^m$, for any posterior distribution $\mathcal Q $ such that $\mathcal Q\ll \mathcal P$ and $\mathcal P \ll \mathcal Q$,
\begin{equation}\label{eq:ilezhang}
    \expect{f\sim\mathcal Q}\mathcal E(f)\leq \frac{ae}{e-1}\left(1-\exp\left(
            -\frac{1}{a}\expect{f\sim Q}\mathcal E_m(f) - \frac{\mathcal K(Q, \mathcal P) + \log\frac{1}{\delta}}{m}
\right)\right).
\end{equation} 
\end{theorem}
The proof of \cref{thm:ilezhang} is inspired by the proof of Theorem 10.4 in \citep{giguere_2013}. We first prove an exponential bound on the identity, with a slack parameter $a$. \cref{fig:id-exp-bound} illustrates the bound.

\begin{prop}
[Exponential bound on the identity, with slack parameter]
\label{prop:idexptemp}
For any $a > 1$, 
\begin{equation}
\forall x\in[0,1], ~ x \leq \frac{ae}{e-1}\left(1 - e^{-\frac{x}{a}}\right).
\end{equation}
\end{prop}
\begin{proof}
    The function $f$ defined by $f(x) = \frac{e}{e-1}(1-e^{-x})$ on $[0,1]$ is concave. Since, $f(0)=0$ and $f(1)=1$, $\forall x\in[0,1], ~ x \leq f(x)$ ($f$ is over its chords). The result follows because $\frac{1}{a}[0,1]\subset[0,1]$ ($a>1$).
\end{proof}
\begin{proof}(\cref{thm:ilezhang})
Let $\mathcal P$ be a prior distribution on regression functions. Let $\delta > 0$ and $Q$ be a distribution over regression functions.

Let $a > 1$. Denote: $\zeta(f(x), y) = \frac{\Delta(f(x), y)}{a}$. Then, by \cref{thm:zhang}, with probability at least $1-\delta$ on the training set sampled iid from $\rho^m$, we have

\begin{equation*}
-\expect{g\sim Q}\log\expect{x,y \sim D}e^{-\zeta(g(x), y)} \leq \frac{1}{m}\left(
\expect{g\sim Q}\sum_{i=1}^m \zeta(g(x_i), y_i) + \mathcal K(Q, \mathcal P) + \log\frac{1}{\delta}
\right).
\end{equation*} 
Because $x\mapsto -\log x$ is convex, by Jensen,
\begin{equation*}
-\log\expect{g\sim Q}\expect{x,y \sim D}e^{-\zeta(g(x), y)} \leq \frac{1}{m}\left(
\expect{g\sim Q}\sum_{i=1}^m \zeta(g(x_i), y_i) + \mathcal K(Q, \mathcal P) + \log\frac{1}{\delta}
\right).
\end{equation*} 
Because $x\mapsto -\exp(-x)$ is non-decreasing,
\begin{equation}\label{eq:pb-step1}
    -\expect{g\sim Q}\expect{x,y \sim D}e^{-\zeta(g(x), y)} \leq -\exp\left(-\frac{1}{m}\left(
\expect{g\sim Q}\sum_{i=1}^m \zeta(g(x_i), y_i) + \mathcal K(Q, \mathcal P) + \log\frac{1}{\delta}
\right)\right).
\end{equation} 
On the other hand, from \cref{prop:idexptemp} applied to $\Delta(f(x), y)$, 
\begin{equation}\label{eq:pb-step2}
\Delta(f(x), y) \leq \frac{ae}{e-1}\left(1 - e^{-\Delta(f(x), y)/a}\right).
\end{equation}
We conclude by combining \cref{eq:pb-step1} and \cref{eq:pb-step2}.
\end{proof}

It is important to note that the value of the bound depends on the performance of a given distribution of predictors on the training set, independently of how the distribution is obtained. We return to this discussion in \cref{sec:discussion}.

\section{PAC-Bayes analysis of consistency}\label{sec:consistency}
In the previous section, we proved a PAC-Bayes bound on the expected risk of the stochastic predictor $\mathbb E \Delta(f(x), y)$ (the expectation is taken with respect to $f \sim \mathcal Q$ and $x, y \sim \rho$). The bound depends on the empirical risk of the stochastic predictor $f$ $\mathbb E\frac{1}{m}\sum_{i}\Delta(f(x_i), y_i)$ (where the expectation is taken with respect to $f\sim\mathcal Q$ only) and a penalty term. This can give a tight estimate of the generalization properties of a supervised classification problem. 

However, minimizing the upper bound is at least as hard as solving the ILE Structured prediction problem directly. Thus, it does not provide an efficient way of learning: we cannot, given a model of posterior, estimate the parameters for which we minimize the bound. Furthermore, the results introduced in \cref{sec:pbclass} pertain to generalization and do not quantify the sub-optimality of the predictor of $f$ (compared to $f^*$) as a function of the sub-optimality of $g$.

We propose an approach to solving the latter problem in this section. In \cref{sec:algorithms} we use these results to learn a posterior $\mathcal Q$ over predictors from a training set $(x_1, y_1), \ldots, (x_m, y_m)$.

In this section, we aim to bound the expected excess risk of a (stochastic) predictor $f$ using the empirical excess quadratic surrogate risk. Our contributions are original in several respects: first, they incorporate the finite-sample performance of the predictor distribution in true PAC-Bayes fashion, giving actionable insight into the value of a predictor distribution; second, in the case of finite-dimensional vector spaces, they formalize certain trade-offs that
are important in the practice of Structured prediction.

\subsection{Augmented learning problem}
To this end, we introduce an \emph{augmented learning problem}. Simply put, we consider the learning problem of learning from the dataset $\left(x_i, g^*(x_i)\right)$ instead of learning from $\left(x_i, y_i\right)$. Note that $g^*$ depends on the data-generating distribution $\rho$ but not on the iid samples $x_1, \dots, x_m$. Thus, the pairs $(x_i, g^*(x_i))$ are iid, just like $(x_i, y_i)$.

This allows us to absorb the intractable $\mathcal R(g^*)$ in the Comparison inequality, and decompose $\mathcal R(g) - \mathcal R(g^*)$, using the Strong Comparison Inequality. We introduce the absolute deviation loss $\norm{g(x) - g^*(x)}$, naturally interpretable as the pointwise surrogate excess risk. The empirical sum of these quantities can be seen as the $L_{2,1}$ norm of the residuals of the augmented regression. 

Proving a PAC-Bayes bound for this augmented learning problem requires establishing a concentration inequality over the absolute deviation regression problem (an unbounded loss). In the following section, we present and adapt a result from~\cite{haddouche2020} and establish our result.

\subsection{Absolute regression bound}
In this section, we prove \cref{thm:ilelinear}, which provides a finite-sample dependent consistency guarantee with high-probability. We first state \cref{thm:ilelinear}, then introduce and extend results from \cite{haddouche2020} needed to prove it.

In order to handle the unbounded absolute deviation loss, recall the following definition from \cite{haddouche2020}:

\begin{definition}[Hypothesis-dependent range (\texttt{HYPE}) condition]
    Let $\mathcal D = \mathcal X \times \mathcal H$ be the data space and $\mathcal G$ be the space of predictors.
    A loss function $\ell: \mathcal H \times \mathcal G$ verifies the Hypothesis-dependent range (\texttt{HYPE}) condition for a function $K: \mathcal G \rightarrow \mathbb R^*_+$ such that $\forall g\in\mathcal G, \sup_{x,y \in\mathcal D}\ell(g, z) \leq K(g)$. In this case, we say that $\ell$ is $\texttt{HYPE}(K)$-compliant.
\end{definition}

The following theorem is extended to vector-valued regression from \citet[Thm. 5.1.]{haddouche2020}.

\begin{theorem}[extended from \citealp{haddouche2020}, Thm. 6.1]\label{thm:haddouchelin}
    Let $1 \geq \alpha > 0$ and $N \geq 6$, the dimension of the regression space $\mathcal G$. We define $\ell$ as:
    \begin{equation}
        \ell: (h, (X(x), g^*(x))) \in \mathcal G \times (\mathcal F \times \mathcal H) \mapsto \norm{g^*(x) - h\circ X(x)}.
    \end{equation}
    Then, $\ell$ is $\texttt{HYPE}(K)$ compliant for $K(h)=B\norm{h}_F+C$ with $C = \norm{g^*}_{2, \infty}$ and $B=\norm{X}_{2, \infty}$.

Moreover, let $\mathcal P=\mathcal N(0, \sigma^2I_N)$ with $\sigma^2= t\frac{m^{1-2\alpha}}{B^2}$ with $0 < t <1$, be a Gaussian prior. We have with probability at least $1-\delta$ over $S\sim D^m$, for any posterior distribution $\mathcal Q$ such that $Q\ll P$ and $P\ll Q$:
\begin{align}
\expect{h\sim Q}R(h) \leq& \expect{h\sim Q}R_m(h) + \frac{\mathcal K (Q, P) + \log(2/\delta)}{m^\alpha} + \frac{C^2}{2m^{1-\alpha}}\left(1+F(t)^{-1}\right)\\
                         & +\frac{N}{m^\alpha}\left[\log\left(1+\frac{C}{\sqrt{2F(t)m^{1-2\alpha}}}\right) + \log\left(\frac{1}{\sqrt{1-t}}\right)\right].
\end{align}
where $F(t) = \frac{1-t}{t}$. 
\end{theorem}

\begin{proof}
    We first prove that $\ell$ is $\texttt{HYPE}(K)$-compliant for $K(h)$ of the form $K(h)=\norm{g^*}_{2, \infty} + \norm{h}_{op}\norm{X}_{2, \infty}$. Indeed, 
    \begin{equation*}
        \norm{g^*(x) - h\circ X(x)} \leq \norm{g^*}_{2, \infty} + \norm{h}_{op}\norm{X(x)}_2.
    \end{equation*}
    Furthermore, we have $\forall h, \norm{h}_{op} \leq \norm{h}_F$. Indeed, let $h = U\Sigma V^T$ be the singular value decomposition of $h$ where $U$ and $V$ are unitary and $\Sigma$ is diagonal with non-negative $\sigma_i$ singular values of $A$. Because $U$ and $V$ are unitary, $\norm{\Sigma}_F=\norm{U^ThV}_F = \norm{h}_F$. Thus, 
    
    \begin{equation*}
        \norm{h}_F = \norm{\Sigma}_F = \sqrt{\sum_{i} \sigma_i(A)^2} \geq \max_i \sigma_i(A) = \norm{h}_{op}.
    \end{equation*}
    Thus, 
    
    \begin{equation*}
        \norm{g^*(x) - h\circ X(x)} \leq \norm{g^*}_{2, \infty} + \norm{h}_{op}\norm{X}_{2, \infty}.
    \end{equation*}

The second part of the result is a generalization of the original result. Because the \texttt{HYPE} condition is verified with the same form as in the scalar case. See \citet[Appendix F.2.]{haddouche2020} for a proof of the original result.
\end{proof}

Finally, we specialize \cref{thm:haddouchelin} into a PAC-Bayes generalization bound for the augmented regression problem:
\begin{theorem}
[Augmented linear regression excess risk bound]\label{thm:ilelinear}
Let $\alpha > 0$ and $\dim(\mathcal H\otimes\mathcal F)=N \geq 6$. Let $\mathcal P=\mathcal N(0, \sigma_0^2I_N)$ with $\sigma_0^2= t\frac{m^{1-2\alpha}}{\kappa^2}=t\sigma^2$ be a Gaussian prior with $0 < t <1$. We have with probability at least $1-\delta$ over the training set sampled from $\rho^m$, for any posterior distribution $\mathcal Q=\mathcal N(W, \sigma'^2I_N)$,
\begin{equation}
{\expect{f\sim \mathcal Q}\mathcal E(f) - \mathcal E(f^*)}\leq 2c_\Delta\left[\expect{
g\sim \mathcal Q}\frac{1}{m}\sum_{i=1}^m\norm{g^*(x_i) - g(x_i)} + \frac{\mathcal K \left(\mathcal Q, \mathcal P\right) + \log\frac{2}{\delta}}{m^\alpha} + \varepsilon\left(m, t, \alpha,\mathcal P\right)\right],
\end{equation}
where
\begin{equation}
\varepsilon\left(m, t, \alpha, \mathcal P\right) = \frac{\norm{g^*}^2}{2m^{1-\alpha}}\left(1+F(t)^{-1}\right) +\frac{N}{m^\alpha}\left[\log\left(1+\frac{\norm{g^*}}{\sqrt{2F(t)m^{1-2\alpha}}}\right) + \log\left(\frac{1}{\sqrt{1-t}}\right)\right],
\end{equation}
and $F(t) = \frac{1-t}{t}$. 
\end{theorem}

\begin{proof}
    Recall the  Strong Comparison Inequality (\cref{thm:strongcomp}):
    \begin{equation*}
        \mathcal E(f) - \mathcal E(f^*) \leq 2c_\Delta\underbrace{\int_\mathcal X \norm{g(x)-g^*(x)}_\mathcal H d\rho_\mathcal X (x)}_\text{$=: R(g)$}.
    \end{equation*}

    We can then apply the Augmented regression formulation and \cref{thm:haddouchelin} to bound the expectation of the lower-bound, by recognizing $R(g)$. We also specialize the result for Gaussian posteriors.
\end{proof}

\cref{thm:ilelinear} expands on the general PAC-Bayes approach to learning presented in \cref{sec:pbclass} and on the Comparison Inequality \cref{thm:comp} by considering the consistency problem from the point of view of finite sample concentration: how suboptimal is $f$ with respect to $f^*$ over the data-generating distribution given the suboptimality of $g$ with respect to $g^*$ on the finite, real-world dataset at hand. 

In the rest of this section we focus on the analysis of the dependencies of the bound in \cref{thm:ilelinear} in the different learning problem and bound parameters. By examining and discussing the influence of these parameters on the value of the bound, we show that our approach yields practical guidance on the performance of learning methods and formulations, in particular on the choice of loss for Structured prediction. 

\subsection{Posterior parametrization}\label{sec:kl}

We first examine the impact of the choice of posterior variance on the Kullback-Leibler divergence term. Different parametrizations yield different regularization behaviors, we present two alternatives here.

For completeness, we begin by recalling the Kullback-Leibler divergence of two multivariate Gaussian distributions, with a specialization for isotropic Gaussians.

\begin{prop}
[Kullback-Leibler divergence of multivariate Gaussians]
The Kullback-Leibler divergence of two $N$-dimensional multivariate Gaussians $\mathcal P = \mathcal N(\mu_2, \Sigma_2)$ and $\mathcal Q = \mathcal N(\mu_1, \Sigma_1)$ is:
\begin{equation}
\label{eq:kl}\mathcal K(\mathcal Q, \mathcal P) = \frac{1}{2}\left[
\log{\frac{\det{\Sigma_2}}{\det{\Sigma_1}}}
-N
+tr(\Sigma_2^{-1} \Sigma_1)
+(\mu_2 - \mu_1)^T\Sigma_2^{-1}(\mu_2-\mu_1)
\right].
\end{equation}
In particular, if $\mu_2=0$, $\Sigma_2 = \sigma_2^2I_N$ and $\Sigma_1=\sigma_1^2I_N$, then \cref{eq:kl} can be reduced to:
\begin{equation}
\mathcal K(\mathcal Q, \mathcal P) = \frac{1}{2}\left[
2N\log{\frac{\sigma_2}{\sigma_1}}
-N
+N\frac{\sigma_1^2}{\sigma_2^2}
+\frac{\norm{\mu_1}^2}{\sigma_2^2}
\right].
\end{equation}
\end{prop}

The proof can be found in many references, including in \cite{matrix-cookbook}.

\begin{figure}[t]
\begin{center}
     \includegraphics[width=0.8\textwidth]{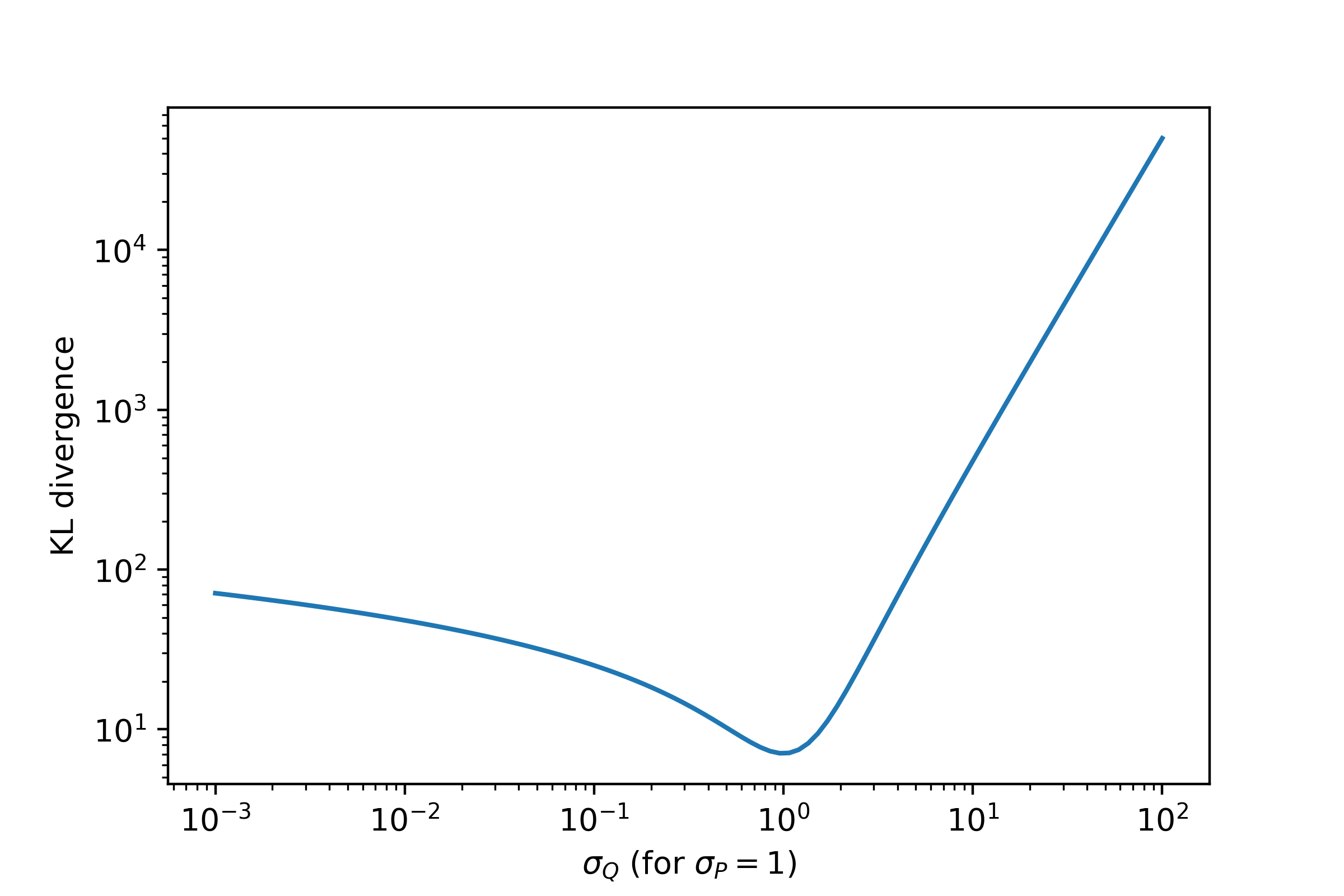}
\end{center}
\caption{Kullback-Leibler divergence as a function of posterior covariance $\sigma$, where the prior covariance is unitary. Note the global minimum.}
\label{fig:parametrization}
\end{figure}

\paragraph{General Kullback-Leibler variations.} For ease of comprehension, we consider the general case where the variance of $\mathcal P$ is fixed, and study the variations of the Kullback-Leibler divergence as the variance of $\mathcal Q$ changes. In \cref{fig:parametrization}, we illustrate the dependence of the Kullback-Leilber divergence on posterior variance (especially relatively to the prior variance).

The main takeaway of such an illustration is that there is a global minimum of the Kullback-Leibler divergence (even though the means do not coincide), which justifies that the ensuing discussion in the section is justified.

Of course, the posterior variance can be optimized in all generality. However, simple rules are interesting because they can be applied in practice. Furthermore, in PAC-Bayes literature, it is common for the posterior variance to be fixed \emph{a priori}, for example to $1$. Two \emph{a priori} examples are interesting because they yield insight into the impact of the problem parameters on generalization performance: the unit-variance model and the so-called, wide posterior. We first
present both models then compare them.

\paragraph{Unit variance posterior} This posterior model is widely used in PAC-Bayes analysis. Here, $\sigma_1^2=1$ and $\sigma_2^2=t\sigma^2$ where $0 < t < 1$, as defined in \cref{thm:ilelinear}. Denote $\mathcal K_U(t)$ the Kullback-Leibler divergence in this model, which has the following expression:
\begin{equation}
\mathcal K_U(t)= \mathcal K(\mathcal Q, \mathcal P) = 
\frac{\norm{\mu_1}^2}{2t\sigma^2}
+ \frac{n}{2}\left(
\log{t}
+\log{\sigma^2}
-1
+\frac{1}{t\sigma^2}
\right).
\end{equation}
\paragraph{Wide posterior} In this case, we set the posterior variance to the supremum of the set of admissible variances for the prior (i.e. corresponding to $t=1$ in \cref{thm:ilelinear}). Here, $\sigma_1^2=\sigma^2$ and $\sigma_2^2=t\sigma^2$, where $0 < t < 1$, and the Kullback-L divergence, noted $\mathcal K_W(t)$ becomes:
\begin{equation}
\mathcal K_W(t)=\mathcal K(\mathcal Q, \mathcal P) = 
\frac{\norm{\mu_1}^2}{2t\sigma^2}
+\frac{n}{2}\left(
\log{t}
-1
+\frac{1}{t}
\right).
\end{equation}

\paragraph{Comparison} We compare the two parametrizations, to better understand the implications of each hypothesis. The difference of the two divergences gives:
\begin{equation}
\mathcal K_U(t) - \mathcal K_W(t)
= \frac{n}{2} \left(
\log\sigma^2 + \frac{1}{t\sigma^2} - \frac{1}{t}
\right).
\end{equation}

This gap depends on $t$ but only through the prior variance. It is positive if and only if 
\begin{equation}
\log\sigma^2 + \frac{1}{t\sigma^2} - \frac{1}{t} > 0.
\end{equation}
If $\sigma^2 < 1$, this is always verified. If $\sigma^2 > 1$, then there is a threshold behavior at $t_0(\sigma) = \frac{1 - \frac{1}{\sigma^2}}{\log\sigma^2}$.

In \cref{thm:ilelinear}, $\sigma$ is not arbitrary, but determined by the parameters of the learning problem. Recall that $\sigma = \frac{m^{1-2\alpha}}{\kappa}$ where $\kappa^2 = \sup_{x\in\mathcal X}k(x, x)$. In other words, the magnitude of $\sigma$ is inversely proportional to that of $\kappa$, with coefficient $m^{1-2\alpha}$.

In the special case of $\alpha=1/2$, the above discussion yields insight into the impact of the magnitude of the input kernel on the generalization performance of the ILE method. 
\begin{itemize}
    \item If the input kernel is bounded by $1$, then $\sigma > 1$ and the choice of parametrization (and the choice of $t$ has an impact). In this case, $t$ should be chosen larger than the threshold value when the wide parametrization is chosen, and smaller when the unitary parametrization is selected. This remark nuances the remark in \cite{haddouche2020} that the limitation on the choice of prior is not limiting: choosing a smaller prior variance helps obtain good generalization bounds with a posterior with smaller variance.
    \item If the input kernel cannot be bounded by $1$, then $\sigma < 1$ and the gap is always positive. In this case, the wide parametrization is better, whichever the choice of $t$. Note however that if the known bound for a given kernel is not tight, then the practitioner may find herself in the previous case without knowing it. This highlights the importance of having informative bounds when controlling the different ingredients to the learning problem.
\end{itemize}

The analysis in more subtle with $\alpha \neq \frac{1}{2}$, because the prior variance depends on $m$. 

\begin{itemize}
    \item In the case where $\alpha < \frac{1}{2}$, $\sigma \rightarrow + \infty$ when $m \rightarrow \infty$. So with enough data, we are in the first case above. Note that as $\sigma$ becomes large, the threshold value above becomes small, meaning that the potential widening of the posterior is compensated by the choice of a small $t$ (in the unitary posterior parametrization).
    \item If $1 > \alpha > \frac{1}{2}$, the prior becomes tighter as the quantity of data grows. This seems pathological in the large data limit: we want to maintain variance in the prior as the quantity of data grows.
\end{itemize}

The above discussion gives insight into the choice of posterior parametrization, giving simple rules of thumb for practical use (which eliminates a free parameter). Furthermore, we have shown that having informative bounds on the different ingredients of the learning problem has impact of the generalization guarantees we can obtain.

In the rest of this section, we qualitatively study the effect of different bound parameters on the bound value. We consider the unit-variance posterior parametrization. Furthermore, to evaluate the effects of the parameters on the bound values, we consider the mean regressor $\bar g = g^*$. The discussion can be formulated in the following way: 

\begin{quote}
    If $g^*$ was known, what PAC-Bayes guarantees could we expect on the resulting stochastic predictor?
\end{quote}

In the rest of this section, we adopt the following shorthand, where we have regrouped terms in $\varepsilon^\prime = \frac{\mathcal K(\mathcal Q, \mathcal P)}{m^\alpha} + \varepsilon$ to highlight their effect on the bound:
\begin{align}
\varepsilon^\prime = \frac{N}{m^\alpha}\left[\log\left(1+\frac{\norm{g^*}}{\sqrt{2F(t)m^{1-2\alpha}}}\right) + \log\left(\sqrt{F(t)^{-1}}\right)  + \frac{1}{2}\log\left(\frac{m^{1-2\alpha}}{\kappa^2}\right) -\frac{1}{2}\right]\nonumber\\
+\frac{N}{2m^{1-\alpha}}\left[\frac{\kappa^2}{t}\left(1 + \frac{\norm{g^*}^2_2}{N}\right) +\frac{\norm{g^*}^2_2}{N}\left(1+F(t)^{-1}\right)\right]\label{eq:penalty}.
\end{align}

\subsection{Penalty dependence on $N$} 
The bound's dependence on $N$, the dimension of the regressor space, is a key contribution of our work. It justifies the practitioner's intuition in the choice of losses for Structured output prediction is important.

We can see in \cref{eq:penalty} that $N$ contributes greatly to the generalization penalty, in $O(N)$. However, as written \cref{eq:penalty} does not materialize the dependence of the penalty on the ``size'' of the Structured output learning problem. We discuss below how the choice of loss determines the penalty value through $N$. Recall that $N = \dim(\mathcal H) \times \dim(\mathcal F)$ is the dimension of the regressor. Indeed, the choice of loss impacts the ILE which in turn impacts the dimension of $\mathcal H$.

To illustrate this effect concretely, let us return to the multi-label binary classification problem in \cref{ex:example}. For the $0-1$ loss, $\varphi(y)$ is a one-hot encoding of $y$, and $\dim(\mathcal H) = \vert \mathcal Y \vert
= 2^\ell$. For the Hamming loss, $\dim(\mathcal H) = 2\ell + 1 \ll \vert\mathcal Y\vert$. This greatly reduces the number of samples needed to attain good generalization performance and, by \cref{thm:ilelinear}, consistency.

In other words, the choice of a ``structured loss'' harnesses the power of the framework described above, instead of considering the Structured prediction problem as naïve classification.

\subsection{Penalty dependence over $m$} \cref{eq:penalty} shows that apart from a light logarithmic dependence in $m$, the penalty decreases as $O\left(\frac{1}{m^\beta}\right)$, where $\beta = \min\left(\alpha, 1-\alpha\right)$. This gives an indication of a rate of convergence, which we discuss more generally in \cref{sec:discussion}, and echoes that in \cref{sec:kl}.

\subsection{Penalty dependence over $t$} 
\begin{figure}[t]
\begin{center}
    \includegraphics[width=0.8\textwidth]{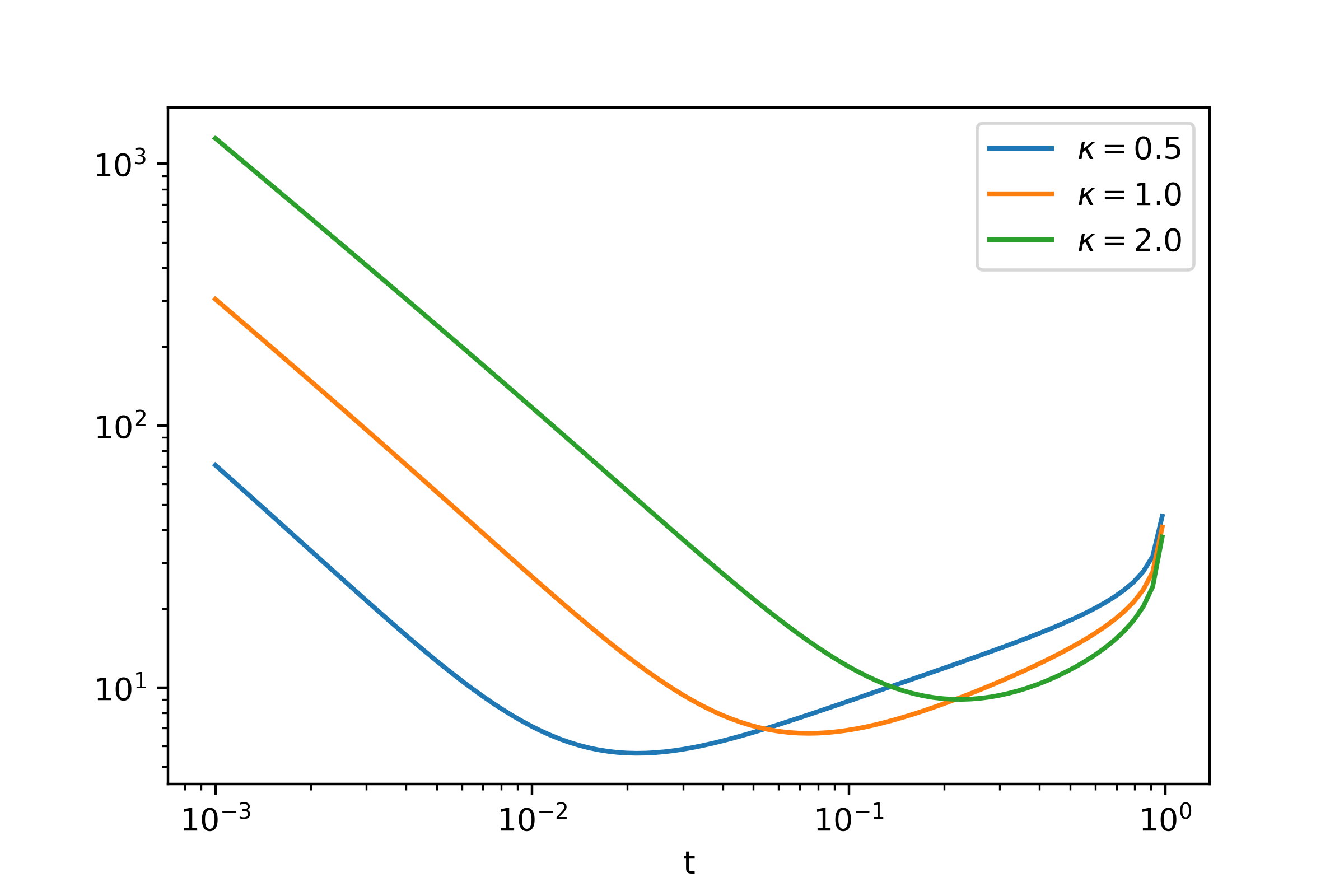}
\end{center}
\caption{Dependence of $\varepsilon^\prime$ as a function of $t$, for different values of $\kappa$. For these curves: $N=10^2$, $m=10^4$, $\alpha=0.3$ and $\norm{g^*}=10$. Note the global minimum over $0 < t < 1$.}
\label{fig:bound-t}
\end{figure}

In \cref{fig:bound-t}, we visualize the dependence of \cref{eq:penalty} on $t$ for different values of $\kappa$. Note the global minimum over $t$.

\section{Learning from the Augmented Regression Bound}\label{sec:algorithms}
In this section, we present and analyze learning algorithms derived from \cref{thm:ilelinear}. In PAC-Bayes theory, learning algorithms are derived from a bound when the bound is optimized with respect to the posterior for a given set of data. Because we use a fixed-variance posterior model, we can reduce optimizing the bound in \cref{thm:ilelinear} to minimizing the following objective function, where $\mathcal Q(\bar g)$ is the parameterized distribution with mean $\bar g$:
\begin{equation}
    J(\bar g) = \expect{g\sim \mathcal Q(\bar g)}\frac{1}{m}\sum_{i=1}^m\norm{g^*(x_i) - g(x_i)} + \frac{\norm{\bar h}^2}{2\sigma_0^2m^\alpha}\label{eq:optlintrue}.
\end{equation}

Because $g^*$ is unknown, the above objective is intractable. By applying the Triangle Inequality $\norm{g^*(x) - g(x) } \leq \norm{g^*(x) - \varphi(y)} + \norm{\varphi(y) - g(x)}$, we separate the approximation and estimation errors respectively and can concentrate on the latter. 

Thus, we consider the optimization objective in \cref{eq:optlin} instead of in \cref{eq:optlintrue}:
\begin{equation}
\hat J(\bar g) = \expect{g\sim \mathcal Q(\bar g)}\frac{1}{m}\sum_{i=1}^m\norm{\varphi(y_i) - g(x_i)} + \frac{\norm{\bar h}^2}{2\sigma_0^2m^\alpha}\label{eq:optlin}.
\end{equation}

Although $\hat J$ is convex, because a practical closed-form expression of $\hat J$ does not exist\footnote{Computing the above expectation reduces to computing the mean of the norm of an isotropic Gaussian vector. The norm of a Gaussian vector follows a Generalized Rayleigh Distribution (see for example, \cite{rayleigh_moments} or \cite{rayleigh_properties}). \cite{rayleigh_moments} shows that there exists a closed-form expression of the moments of the Generalized Rayleigh distribution. However,
these expressions depend on the confluent hypergeometric function, which is not closed for in our setting: it is worth noting that in \cite{rayleigh_moments} the words \emph{closed-form} are in quotes!}, we present two approximate approaches: a deterministic relaxation strategy in \cref{sec:relaxation}, then an estimation strategy using the so-called ``log-prob'' trick, and apply variance reduction in \cref{sec:ssgd}.

\subsection{Relaxation strategy}\label{sec:relaxation}
We can upper-bound the $\hat J$ with an expectation-free convex objective $\hat J_c$. In this section, after establishing the upper-bound, we discuss optimizing $\hat J_c$.

\subsubsection{Relaxation}
In order to prove the relaxation in \cref{eq:ilelinobj} we first bound the expected value of the norm of the regression residual.
\begin{prop}
    [Expected deviation upper-bound]\label{prop:relaxation}
Let $\mathcal Q = \mathcal N(W, \sigma^2 I_N)$ a multivariate Gaussian distribution, where (by abuse of notation) $W\in\mathbb R^{d\times d'}$ and $d\times d'=N$. Let $x\in\mathbb R^{d'}$ and $y\in\mathbb R^d$. Then,
\begin{enumerate}
    \item[(a)]$\expect{V\sim \mathcal Q}\norm{y - Vx}^2 = \norm{y-Wx}^2 + \sigma^2d\norm{x}^2$,
    \item[(b)]$\expect{V\sim \mathcal Q}\norm{y - Vx} \leq \sqrt{\sigma^2 d\norm{x}^2 + \norm{y-Wx}^2}$
\end{enumerate}
\end{prop}

\begin{proof}
\textit{(a)} is a classic result: a detailed proof is given for example in \citet[Appendix 10.7.2.]{giguere_2013}.
For \textit{(b)}, we successively apply Jensen's inequality (\cref{eq:step1}) then use \textit{(a)} (\cref{eq:step2}), as detailed below:
\begin{align}
\expect{V\sim Q}\norm{y - Vx} &\leq \sqrt{\expect{V\sim\mathcal Q}\norm{y - Vx}^2},\label{eq:step1}\\
&=\sqrt{\norm{y - Wx}^2 + \sigma^2d \norm{x}^2}.\label{eq:step2}
\end{align} 
\end{proof}

\begin{remark}
    We note that there may in fact several interpretations of (b) in the previous lemma, such as Jensen or the positivity of the variance.
\end{remark}

We can apply this lemma to upper-bound $\hat J(\bar g)$ in \cref{eq:optlin}:
\begin{equation}
\hat J(\bar g)\leq \hat J_c(\bar g) := \frac{1}{m}\sum_{k=1}^m\sqrt{\beta(x_k) + \norm{\varphi(y_k) - \bar h \circ X(x_k)}^2} +\lambda_m^\alpha(t)\norm{\bar h}^2,\label{eq:ilelinobj}
\end{equation} 
where $\beta(x) = \sigma^2\dim(\mathcal H) \norm{X(x)}^2 = \sigma^2\dim(\mathcal H)k(x, x)$ and regularization parameter $\lambda_m^\alpha(t) = \frac{1}{2t\sigma^2 m^\alpha}=\frac{\kappa^2}{2tm^{1-2\alpha}m^\alpha}$.

\begin{remark}[Slackness of the relaxation]
Is the upper-bound in \cref{prop:relaxation} reasonable? A first remark, is that Jensen's inequality is known to be tight. However, we cannot quantify what is lost before training using the bounds (see \cref{sec:experiments}). We evaluated the slackness of \cref{prop:relaxation} (b) numerically in \cref{fig:relaxation-gap}.
\begin{figure}[t]
\begin{center}
    \includegraphics[width=0.8\textwidth]{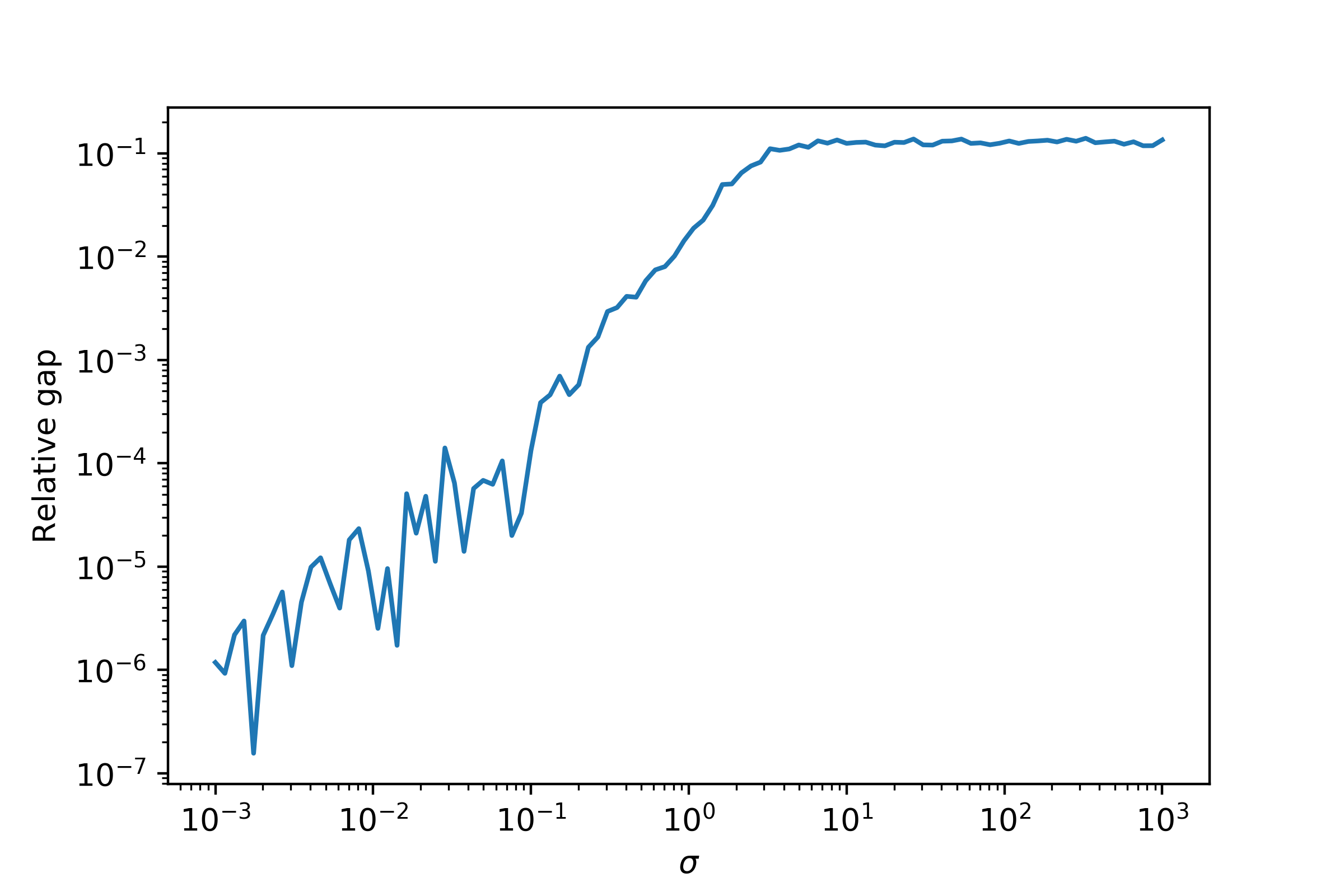}
\end{center}
\caption{Relative gap in \cref{prop:relaxation} (b). We plot $\frac{\vert L(\sigma) - R(\sigma)\vert}{R(\sigma)}$ where $L(\sigma)= \expect{V\sim \mathcal Q}\norm{y - Vx}$ and $R(\sigma) = \sqrt{\sigma^2 d\norm{x}^2 + \norm{y-Wx}^2}$. Notice that the bound is at least tight to $10\%$. The simulation methodology can be found documented at  \github{} in the source code associated to this paper (see \cref{sec:experiments}).}
\label{fig:relaxation-gap}
\end{figure}
\end{remark}

\subsubsection{Learning algorithm}
In this section, we present a minimization method for $\hat J_c(W)$ over $W\in\mathcal L(\mathcal F, \mathcal H)$. The resulting algorithm, based on gradient descent, is presented in \Cref{alg:gd-relax}.

\begin{prop}[Properties of $\hat J_c$]\label{prop:grad-relaxation}
$\hat J_c$ is differentiable and $\mu$-strongly convex for any $\mu>2\lambda_m^\alpha(t)$. Furthermore, $\grad\hat J_c$ is given by
\begin{equation*}
    \grad\hat J_c(W) = \frac{1}{m}\sum_{k=1}^m \frac{\left(\varphi(y_k) - WX(x_k)\right)X(x_k)^T}{\sqrt{\beta(x_k) + \norm{\varphi(y_k) - W X(x_k)}^2}} +2 \lambda_m^\alpha(t)W,
\end{equation*}
thus $\hat J_c$ is twice-differentiable.
\end{prop}

\begin{proof}
Let $l(x) = \sqrt{a + x^2}$ where $a >0$. $l$ is increasing and convex: $l^\prime(x) = \frac{2x}{\sqrt{a+x^2}} > 0 $ and $l^{\prime\prime}(x) = \frac{4a}{2(a+x^2)^{3/2}} > 0$. Furthermore, $W \mapsto \norm{y-Wx}$ is convex. Thus, $W \mapsto \sqrt{\beta(x) + \norm{y-Wx}^2}$ is convex. The result follows by convex combination. Because the regularization penalty is $\mu$ strongly convex for any $\mu > 2 \lambda_m^\alpha(t)$.

Furthermore, if  $f(W) = \norm{Y - WX}^2$, for small $H$,
\begin{align*}
    f(W+H) &= \norm{Y^T-X^T(W^T+ H^T)}^2, \\
        &= \norm{Y^T-X^TW^T}^2 + \norm{X^TH^T}^2 - 2\langle Y^T - X^TW^T, X^TH^T\rangle, \\
        &= \norm{Y-X^TW^T}^2 + \norm{X^TH^T}^2 - 2\Tr[(Y^T - X^TW^T)^TX^TH^T],\\
        &= f(W) - 2\Tr[(Y - WX)X^TH^T] + o(H),\\
        &= f(W) - 2\Tr[X(Y - WX)^TH] + o(H),
\end{align*} 
where we used $\Tr(AB)=\Tr(BA)$ and $\Tr(A^T) = \Tr(A)$. The gradient expression follows by composition.
\end{proof}

Because $\hat J_c$ is $L$-smooth (for some $L>0$), it is well-known that we can minimize it by gradient descent \cite{boyd}. 
\begin{algorithm}[t]
\SetAlgoLined
\SetKwInOut{Input}{Input}
\SetKwInOut{Parameter}{Parameters}
\Input{Initial posterior mean $W_0$.}
\Parameter{Step-size parameters: $\nu\in(0.5, 1]$, $w \geq 0$.}
Initialization $W^0 \leftarrow W_0$.\\
Pre-compute $\beta(x_k)$.\\
\While{\texttt{stopping criterion not met}}{
    Compute $\nabla J_c(W_t)$ (see \cref{prop:grad-relaxation}).\\
Compute step-size: $\gamma_t = \frac{1}{(w + t)^\nu}$.\\
Gradient step: $W^{t+1} \leftarrow W^t - \gamma_t \nabla J_c(W_t)$.
}
\KwResult{$W^{final}$}
\caption{Gradient descent, \texttt{Relax-GD}}
\label{alg:gd-relax}
\end{algorithm}

 \subsection{Stochastic Search Variational Minimization}\label{sec:ssgd}
 In this section, we present a second approach to optimizing $\hat J$. Instead of relaxing the optimization problem, we seek to directly estimate $\grad \hat J$. The intuition behind this approach is that $\hat J$ is convex, thus estimating its gradient could be sufficient to minimize it in practice. To this end, we apply the well-known ``log-prob'' trick. In order to reduce the variance of the resulting estimator, we apply techniques presented in \cite{rosca} and \cite{paisley} with a well-chosen control variate.

 These techniques are widely-used in different fields of mathematics and in particular machine learning: stochastic optimization, reinforcement learning, extreme-value statistics, among others. The interested reader can consult the recent survey \cite{rosca} and references therein.

 \subsubsection{Gradient estimation}
 Without loss of generality, we ignore the quadratic regularization in $\hat J$ and focus of the expectation term, using the following notation:
 \begin{equation*}
     \hat J(W)  = \expect{\mathcal Q(V\vert W)}\left[ L(V)\right],
 \end{equation*}

 where $L(V)= \frac{1}{m}\sum_{k=1}^m\norm{\varphi(y_k) - VX(x_k)}$ and $\mathcal Q(V\vert W)$ is the density of $\mathcal Q(W)$ at $V$.

 The following proposition provides an expression of $\nabla \hat J$ at $W$ and derives an unbiased estimator of $\nabla \hat J$.

 \begin{prop}[Score-function gradient expression]\label{prop:sf-grad}
     $\nabla \hat J(W)$ can be expressed as the expectation of a function of $V\sim\mathcal Q(W)$ over $\mathcal Q(W)$:
    \begin{equation*}
    \nabla \hat J(W) = \expect{V\sim\mathcal Q(W)}\left[L(V)\nabla\log \mathcal Q(V|W)\right].
    \end{equation*}
\end{prop}
 
\begin{definition}
    [Score function estimator (SFE)]
    Let $V_1, \ldots, V_M$ be $M$ iid copies of $V\sim \mathcal Q(W)$. The score function estimator is defined as:
    \begin{equation}
    \eta_M (W)= \frac{1}{M}\sum_{k=1}^M L(V_k)\nabla\log\mathcal Q(V_k|W).
    \end{equation}
\end{definition}

\begin{prop}\label{prop:sfe-unbiased}
    $\eta_M$ is an unbiased estimator of $\nabla\hat J(W)$.
\end{prop}

The proof of \cref{prop:sfe-unbiased} stems directly from \cref{prop:sf-grad}. Given the importance of the approach in our PAC-Bayes approach, we sketch out the main steps of the proof of \cref{prop:sf-grad}. 

As highlighted in \cite{rosca}, proving \cref{prop:sf-grad} reduces essentially to interchanging the derivative and the expectation, and applying Lebesgue's dominated convergence theorem. Three conditions should be verified:
\begin{enumerate}
    \item $\mathcal Q(v \vert W)$ is continuously differentiable with respect to $W$
    \item $L(v)\mathcal Q(v \vert W)$ is integrable and differentiable with respect to $W$.
    \item $L(v)\nabla \mathcal Q(v\vert W)$ is dominated by an integrable function $g$ of $v$.
\end{enumerate}
\begin{proof}
The first two points are straight-forward for the multivariate Gaussian distribution. It is enough to verify the third over all bounded open sets of the space of linear functions. Let $\mathcal C$ be such a set and $C = \max \lbrace \norm{W}, W\in\mathcal C\rbrace$.

Ignoring the normalization constant of $\mathcal Q$, we have:
\begin{equation}
    \norm{L(V)\nabla \mathcal Q(V\vert W)} = \norm{Y-VX} \norm{V-W} \exp\left(-\frac{1}{2\sigma^2}\norm{V-W}^2\right).
\end{equation}

Noting that $- \norm {V-W}^2 = - \norm{V}^2 - \underbrace{\norm{W}^2}_\text{$\geq 0$} +  2\langle V, W \rangle \leq - \norm{V}^2 +2C\norm{V}$ and $\norm{V-W} \leq \norm{V} + C$, we have:
\begin{equation*}
    \norm{L(V)\nabla \mathcal Q(V\vert W)}\leq \norm{Y-VX}\left(\norm{V} + C\right)\exp\left(-\frac{1}{2\sigma^2}\left[\norm{V}^2 -2C\right]\right) =: \gamma(V).
\end{equation*}

Because $\norm{V}^2\gamma(V) \rightarrow 0$ when $\norm{V} \rightarrow \infty$, $\gamma$ is integrable. 
\end{proof}

\begin{remark}
    In \cref{prop:sf-grad}, $L$ is not required to be differentiable.
\end{remark}

With this estimator of the gradient of $\hat J$, we can then minimize $\hat J$ using gradient descent. We present the associated algorithm in \Cref{alg:sfe_naive}.

\begin{algorithm}[t]
\SetAlgoLined
\SetKwInOut{Input}{Input}
\SetKwInOut{Parameter}{Parameters}
\Input{Initial posterior mean $W_0$.}
\Parameter{Step-size parameters: $\nu\in(0.5, 1]$, $w \geq 0$.}
Initialization $W^0 \leftarrow W_0$.\\
\While{\texttt{stopping criterion not met}}{
Sample $M$ predictors from $\mathcal Q$: $V_1, \ldots, V_M$.\\
Estimate intractable gradient: $\eta_M \leftarrow \frac{1}{M}\sum_k f(V_k)\nabla\log \mathcal Q(V_K\vert W^t)$.\\
Compute step-size: $\gamma_t = \frac{1}{(w + t)^\nu}$.\\
Gradient step: $W^{t+1} \leftarrow W^t - \gamma_t \eta_M$.
}
\KwResult{$W^{final}$}
\caption{Naïve score estimator gradient descent, \texttt{SF-GD}}
\label{alg:sfe_naive}
\end{algorithm}

The algorithm described above does not have the same theoretical guarantees as gradient descent, as the gradient estimate, while unbiased, can have large variance. There are several approaches to controlling the gradient's variance. We present an approach in \cref{sec:variance-reduction}.

\subsubsection{Variance reduction}\label{sec:variance-reduction}
A general method of reducing the variance of an estimator is to introduce a \emph{control variate}.

The main ingredient to this approach is the construction of a function $B(V)$ highly correlated with $L(V)$, so as to guarantee that $B(V)\nabla \log \mathcal Q(V\vert W)$ and $L(V)\nabla\log \mathcal Q(V\vert W)$ will also be correlated.

Let $B(V)$ be such a function, highly correlated with $L$. We define, for $a\in\mathbb R$,
\begin{equation}\label{eq:def-variate}
\tilde L(V):= L(V)\nabla \log \mathcal Q(V) - a\left(B(V)\nabla \log \mathcal Q(V) - \expect{q}B(V)\nabla\log \mathcal Q(V)\right).
\end{equation}

Note that $\expect{\mathcal Q} \tilde L(V) = \expect{\mathcal Q} L(V)\nabla \log \mathcal Q$. The goal is thus to choose $a$ such that the variance of $\tilde L(V)$ is minimized.

\begin{figure}[t]
\begin{center}
    \includegraphics[width=0.8\textwidth]{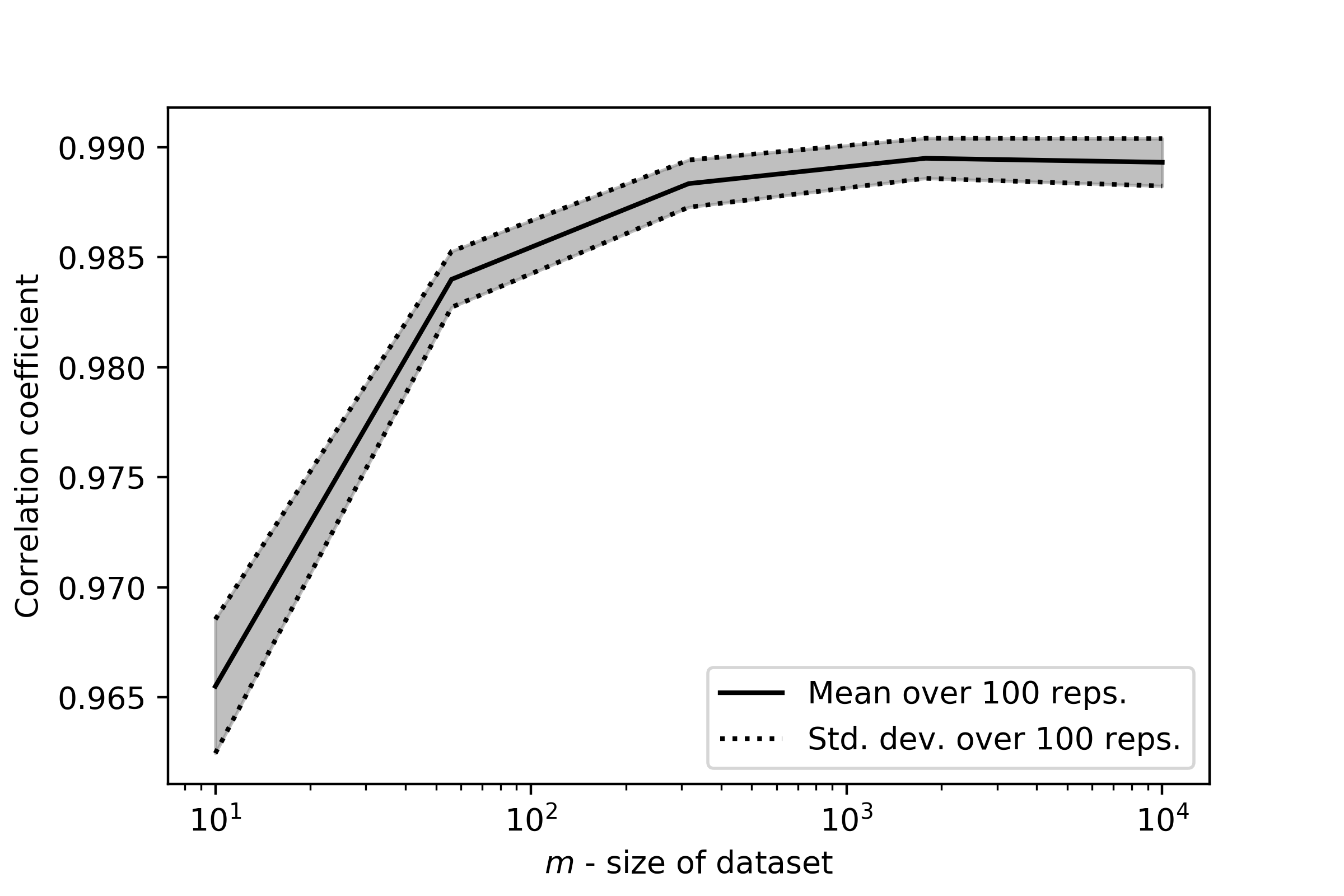}
\end{center}
\caption{Numerical evaluation of correlation for $L(V)=\frac{1}{m}\sum_i^m \norm{y_i - Vx_i}$ and $B(V)=\frac{1}{m}\sum_i^m\norm{y_i-Vx_i}^2$. The correlation coefficient is given over $M=500$ predictors sampled from a isotropic multivariate normal distribution. The coefficients are averaged over $100$ experiments in order to estimate the variance of the estimate (which is materialized by the envelope and dotted lines). To reproduce, see \github.}
\label{fig:correlation}
\end{figure}

\paragraph{Choice of $a$ \& estimation of $a^*$}
\begin{prop}
    [Optimal baseline coefficient, \citealp{paisley}]
    Given $L$ and $B$, define $\tilde L$ as in \cref{eq:def-variate}. Then, $\mathbb V(\tilde L)$ is minimized for $a^*\in\mathbb R$ given by:
    \begin{equation}
        a^* =  \frac{\sum_{k=1}^N\Cov\left(L\frac{\partial\log\mathcal Q}{\partial W_k}, B\frac{\partial\log\mathcal Q}{\partial W_k}\right)}{\sum_{k=1}^N\mathbb V\left(B\frac{\partial\log\mathcal Q}{\partial W_k}\right)}.
    \end{equation}
\end{prop}
Of course, $a^*$ is intractable in general. We use a plug-in estimator of it $\hat a$, by sampling from $\mathcal Q$, independently from the gradient samples. As recommended in 

\begin{definition}
    [Optimal baseline coefficient estimator $\hat a$]

    Let $V^1, \ldots, V^{M^\prime}$ be $M^\prime$ iid samples from $\mathcal Q(V\vert W)$. Then, 
    \begin{equation}
    \label{eq:choice-of-a}
    \hat a = \frac{\alpha}{\beta},
    \end{equation}
   where 
\begin{align}
    \alpha &= \frac{1}{M^\prime}\sum_{j=1}^{M^\prime}\sum_{k=1}^N\Cov\left(L\frac{\partial\log\mathcal Q(V^j)}{\partial W_k}, B\frac{\partial\log\mathcal Q(V^j)}{\partial W_k}\right),\\
    \beta &= \frac{1}{M^\prime}\sum_{j=1}^{M^\prime}\sum_{k=1}^N\mathbb V\left(B\frac{\partial\log\mathcal Q(V^j)}{\partial W_k}\right).
\end{align}
\end{definition}

This is in particular the estimator proposed by \cite{paisley} and \citet[associated code]{rosca}\footnote{See: \href{https://github.com/deepmind/mc_gradients/blob/master/monte_carlo_gradients/control_variates.py}{https://github.com/deepmind/mc\_gradients/blob/master/monte\_carlo\_gradients/control\_variates.py}}.
\paragraph{Choice of $B$} Of course, the variance reduction virtues of this approach rely on having a function $B$ strongly correlated with $L$ and for which we can compute the gradient of its expectation in closed form. \cite{rosca} present several alternative methods for choosing such a $B$. In particular, $B$ does not need to be an upper-bound of $L$ (contrary to the relaxation approach above).

We define $B(V)$ as:
\begin{equation}\label{eq:B-def}
    B(V) = \frac{1}{m}\sum_{i=1}^m\norm{y_i - Vx}^2.
\end{equation}

The intuition behind such a choice is that we have a closed-form expression of the gradient of $B$, and the norm and squared-norm are correlated as one is a non-trivial deterministic function of the other. Of course, $L$ ($B$) are linear combinations of the norms (squared-norms) of residuals (respectively). However, the $(x,y)$ pairs are drawn iid and because $x,y\mapsto \norm{y-Vx}$ is deterministic, $\norm{y - Vx}$ and $\norm{y^\prime - Vx^\prime}$ are
independent. Conditioned on the dataset, the correlation of $L$ and $B$ should remain strong.
\begin{remark}
    [Numerical evaluation of correlation]
    We numerically evaluated our hypothesis that $L$ and $B$ as defined in \cref{eq:B-def} are well correlated, even when the correlation is reduced by summing over the dataset. We present the results in \cref{fig:correlation}. The correlation is strong, including for large datasets. The experiment is detailed in the associated codebase: \github.
\end{remark}

To summarize, the different approximations are as follows (again, ignoring the regularization term): 
\begin{align}
    \nabla \hat J(W) &= \expect{q}\left[L(V)\nabla\log \mathcal Q(V) + \nabla \ell(W)\right],\nonumber\\
                     &= \expect{q}\left[L(V)\nabla \log \mathcal Q(V) - a\left(B(V)\nabla \log \mathcal Q(V) - \grad\expect{q}B(V)\right)\right],\nonumber\\
                     &= \expect{q}\left[(L(V) - aB(V))\nabla\log \mathcal Q(V)\right] +a\grad\expect{q}B(V),\nonumber\\
                     &\approx \underbrace{\expect{q}\left[(L(V) - \hat aB(V))\nabla\log \mathcal Q(V)\right]}_\text{Sample} +\hat a\underbrace{\grad\expect{q}B(V)}_\text{Tractable},\nonumber\\
                     &\approx \underbrace{\frac{1}{M} \sum_{k=1}^M \left(L(V^k) - \hat a B(V^k)\right)\nabla \log \mathcal Q(V^k)}_\text{$\hat \eta_M(W)$} + \hat a \underbrace{\nabla \expect{q}B(V)}_\text{$\eta_B(W)$}.\label{eq:cv-grad-defs}
\end{align}

where $V_1, \ldots, V^M$ are iid and independent from the $M^\prime$ samples used to estimate $\hat a$.

\subsubsection{Algorithm}
We present the complete algorithm using Stochastic Search Gradient Descent \cite{paisley}. To the best of our knowledge, this is a novel application of Stochastic Search Gradient Descent, the applications in the \cite{paisley} being logistic regression and Hierarchical Dirichlet processes. 

As described above, two intractable quantities are sampled in order to estimate the gradient: the stochastic search correction $\hat \eta_M$ and $\hat a$ (as shown in \cref{eq:cv-grad-defs}). Let $M$ and $M^\prime$ be the number of samples used for each estimation, respectively\footnote{Note that \cite{paisley} consider $M^\prime \ll M$.}.

We summarize the Stochastic Search Gradient Descent algorithm with the control variate described above in \Cref{alg:ssgd-full}. It repetitively calls \Cref{alg:ssgd-step}.

\begin{algorithm}[t]
\SetAlgoLined
\SetKwInOut{Input}{Input}
\SetKwInOut{Parameter}{Parameters}
\Input{Previous $W^{t}$.}
\Parameter{Step-size parameters: $\nu\in(0.5, 1]$, $w \geq 0$, $M$, $M^\prime$.}
Sample $M^\prime$ predictors $V_1, \ldots, V_{M^\prime}$ from $\mathcal Q(W^t)$.\\
Estimate $\hat a$ according to \cref{eq:choice-of-a}.\\
Sample $M$ predictors from $\mathcal Q(W^t)$, $V_1, \ldots, V_{M}$.\\
Estimate intractable gradient $\hat \eta_M \leftarrow \frac{1}{M_s}\sum_k (L(V_k) - \hat a B(V_k))\nabla\log \mathcal Q(V_k|W^t)$.\\
Compute $\eta_B = \nabla \expect{} B(V)$ under $\mathcal Q(W^t)$ according to \cref{eq:ggrad}.\\
Compute $\eta_P$ the gradient of the penalty term.\\
Compute step-size $\gamma_t = \frac{1}{(w + t)^\nu}$.\\
Update $W^{t+1} \leftarrow W^{t} - \gamma_t \hat\eta_M(W^t) - \gamma_t \hat a\eta_B(W^t) - \gamma_t\eta_P$.\\
\KwResult{$W^{t+1}$}
\caption{Quadratic Stochastic Search Gradient Descent iteration, \texttt{Q-SSGD}.}
\label{alg:ssgd-step}
\end{algorithm}

\cref{lemma:controlgradient} provide expressions of the closed form gradients that are needed in the algorithm.

\begin{prop}
[Control variate gradient $\eta_B$]\label{lemma:controlgradient}
\begin{equation}\label{eq:ggrad}
\nabla\mathbb E_{\mathcal Q(W)}B(V) = \frac{2}{m}\sum_{k=1}^m \left(\varphi(y_k) - WX(x_k)\right)X(x_k)^T.
\end{equation}
\end{prop}

\begin{proof}
\begin{align*}
    \nabla\expect{}B(V) &= \nabla\left[\frac{1}{m}\sum_{k=1}^m \norm{\varphi(y_k)-WX(x_k)}^2 + \sigma^2\norm{X(x)}^2\dim(\mathcal H)\right],\\
                        &= \frac{1}{m}\sum_{k=1}^m \nabla \left[\norm{\varphi(y_k) - WX(x_k)}^2\right] = \frac{2}{m}\sum_{k=1}^m \left(\varphi(y_k) - WX(x_k)\right)X(x_k)^T.
\end{align*}
\end{proof}

\begin{algorithm}[t]
\SetAlgoLined
\SetKwInOut{Input}{Input}
\SetKwInOut{Parameter}{Parameters}
\Input{Initial regressor $W^{0}$.}
\Parameter{Step-size parameters: $\nu\in(0.5, 1]$, $w \geq 0$, $M$, $M^\prime$.}
\While{\texttt{stopping criterion not met}}{
    $W^{t+1} \leftarrow$ \texttt{Q-SSGD($W^t$)}\\
    $t\leftarrow t+1$
}
\KwResult{$W^{t}$}
\caption{Quadratic Stochastic Search Gradient Descent algorithm.}
\label{alg:ssgd-full}
\end{algorithm}

In this section, we presented two approaches to learning with \cref{thm:ilelinear} -- bringing additional evidence that PAC-Bayes bounds are a principled way to derive new algorithms \citep[see \emph{e.g.} the ICML 2019 tutorial][]{GueSTICML}. In \cref{sec:experiments}, we provide experimental results of to support theoretical results and discussion in \cref{sec:pbclass,sec:consistency,sec:algorithms}.

\section{Discussion}\label{sec:discussion}
In this paper, we provide a theoretical analysis of the properties of estimators in the Implicit Loss Embedding framework. Our analysis is theoretical and methodological in nature. On one hand, we prove PAC-Bayes generalization bounds for ILE losses. These seek to upper-bound the task risk and the excess task risk, from the performance of a given predictor distribution on the dataset. In turn these theoretical results yield two important methodological contributions. 

In this section, we briefly summarize our main contributions at a high-level, tying together the different sections and approaches presented therein. Furthermore, we put our results into perspective with respect to those presented in \cite{ciliberto_general_2020}.

\subsection{Generalization}
In \cref{thm:ilezhang}, we prove a generalization bound for the ILE problem stemming from a classic PAC-Bayes bound from \cite{pb_zhang}. It is important to note that in \cref{thm:ilezhang}, the loss function $\Delta$ is ILE but the bound hold irrespective of how the predictor (distribution) $\mathcal Q$ is obtained. This differs from the approach adopted in \cite{ciliberto_general_2020} which consider convergence rates of specific algorithms.

The bound offers methodological guarantees. In the case where $m$ is very large, the generalization penalty term is small, so with probability $1-\delta$, the generalization gap (i.e. the difference between $\expect{\mathcal Q}\mathcal E(f)$ and $\expect{\mathcal Q}\mathcal E_m(f)$) reduces to the gap induced by exponential bound on the identity (\cref{prop:idexptemp}). For small values of $\expect{\mathcal Q} \mathcal E_m(f)$, this bound is tight (see \cref{fig:id-exp-bound}). This is expected: in the infinite data limit, we expect the empirical loss to be close to the distribution loss. 

Furthermore, generalization performance is limited by a penalization term formed of the Kullback-Leibler divergence of the posterior and prior distributions. In practice, this yields a sound way of penalizing predictor (distributions) that do not generalize well. Indeed, in the simple multivariate normal case, if the distributions have the same variance, the penalization coincides with the widely used Tikhonov regularization.

\subsection{Comparison with KDE generalization bounds}
\cite{ciliberto_general_2020} offers insight into the links between the Kernel Dependency Estimation approach and the ILE framework. In this section, we tie together the PAC-Bayes approach to generalization and the ILE approach to consistency, by recalling some results from \cite{giguere_2013}.

We consider an output space $\mathcal Y$ associated to the reproducing kernel $k_\mathcal Y$ defined by $k_\mathcal Y(y, y^\prime) = \langle Y(y), Y(y^\prime)\rangle, ~\forall y,y^\prime\in\mathcal Y^2$. We assume the task loss is defined from the output kernel $k_\mathcal Y$ as follows:
\begin{equation}
L(y, y^\prime) = \frac{1}{2}\norm{Y(y)-Y(y^\prime)}^2=\frac{1}{2}k_\mathcal Y(y, y) + \frac{1}{2}k_\mathcal Y(y^\prime, y^\prime)  -k_\mathcal Y(y, y^\prime).
\end{equation}
The KDE predictor $f_k$ is the decoded linear regressor, defined by:
\begin{equation}
f_k(x) = \arg\min_{y\in\mathcal Y}\norm{Y(y) - WX(x)}^2.
\end{equation}
One key insight in the KDE approach is to notice that:
\begin{equation}
L(y, f_k(x)) \leq 2\norm{Y(y) - WX(x)}^2\label{eq:kdeupperbound},
\end{equation}
where $W \in \mathcal L(\mathcal F, \mathcal H)$. In this sense, the KDE framework is a quadratic surrogate approach, without any consistency guarantees as is.

\cite{giguere_2013} establishes a PAC-Bayes generalization bound in the KDE setting. The bound is based on the pointwise quadratic upper-bound of the KDE loss we presented in \cref{eq:kdeupperbound} and bounds the expected risk of the mean KDE predictor by the empirical regression risk of the mean regressor, $\mathcal R_m(\bar g) = \frac{1}{m}\sum_{i=1}^{m}\norm{Y(y_i) -
\bar g(x_i)}^2$ and $\bar g = \bar w \circ X$ where $w\in\mathcal L(\mathcal F, \mathcal H)$.

Recall that we assume $\mathcal F$ (the RKHS associated to $k$) and $\mathcal H$ (the Hilbert space associated to the ILE) to be finite-dimensional. We consider regressors $g = w \circ X$ where $w \in \mathcal L(\mathcal F, \mathcal H)$.

\begin{theorem}[\citealp{giguere_2013}]\label{thm:isobound}
Assume $k$ is such that $k(x,x)=1, ~\forall x\in\mathcal X$. For any $\delta > 0$, with probability at least $1-\delta$ over training set sampled iid from $\rho$, for any $w \in \mathcal L(\mathcal F, \mathcal H)$, with $g = w \circ X$,
\begin{equation}
\mathcal E(d \circ g)\leq \frac{5e}{e-1}\left[
1-\exp\left(
- 2\mathcal R_m(g) - \frac{\frac{9}{8}\norm{w}^2 + \log\left(\frac{1}{\delta}\right)}{m}
\right)
\right].
\end{equation}
\end{theorem}

Notice that in the above bound, the predictors are deterministic and not stochastic. Indeed, using Gaussian distributions for $w$, the expectations in \cref{thm:zhang} were explicitly computed, making the bound only depend on the mean predictor associated to $w=\bar w$. 

As \cite{ciliberto_general_2020} point out, under certain hypotheses, we can show that KDE is a special case of ILE and transfer consistency guarantees of ILE predictors to KDE predictors. Thus the surrogate method in the ILE framework can also be seen as the minimization of a PAC-Bayes bound. Even though this is the case, minimizing the upper bound in \cref{thm:isobound} does not yield the same rate guarantees as in \cite{ciliberto_general_2020} which hold for a regularization parameter equal to $\frac{1}{\sqrt{m}}$.

\subsection{Consistency}
Existing approaches do not yield results about the gap between $\mathcal E(d\circ g) - \mathcal E(f^*)$, from a given predictor and a finite sample of data.

Indeed, \cref{thm:learning_rates} provides tight bounds of $\mathcal E(f_m) - \mathcal E(f^*)$ for finite $m$. However, these results depend on how $f_m$ is obtained from the data sample. For example, kernel ridge regression can be used (with regularization parameter $\lambda_m = \frac{1}{\sqrt{m}}$, see \citealp{devito}). For completeness we recall the (abridged) result:

\begin{theorem}
    [\citealp{ciliberto_general_2020}, from Theorem 5]\label{thm:learning_rates}
If $\mathcal Z$ is compact and $\Delta$ admits an ILE. Assume that $k$ is continuous and bounded by $\kappa^2 = \sup_{x\in\mathcal X}{k(x,x)}$. Let $\rho$ be a data-generating distribution such that $g^* \in \mathcal H \otimes \mathcal F$ where $\mathcal H$ is associated to the ILE and $\mathcal F$ to $k$. Let $\delta >0$ and $m_0$ large enough, then with probability at least $ 1 - \delta$ over the training set sampled from $\rho^m$ for $f_m$ (KRR with the same regularization as above),
\begin{equation}
\mathcal E(f_m) - \mathcal E(f^*) \leq \frac{c_\Delta M q \log\frac{4}{\delta}}{m^{1/4}},
\end{equation}
where 
$M = 16\left(\kappa\left(1+\kappa\norm{g^*}\right) + \kappa\sqrt{1 + \norm{g^*}^2} + \norm{g^*}\right)$ and $q \leq 3$.
\end{theorem}

In constrast, in \cref{thm:ilelinear}, we bound a stochastic predictor analog to $\mathcal E(f_m) - \mathcal(f^*)$. This approach is complementary to that of \cite{ciliberto_general_2020}. It provides a theoretical guarantee bounding the suboptimality of a given predictor distribution, and dependent on the finite sample performance of the bound. In particular, \cref{thm:ilelinear} holds with high probability, independently of the learning method used.

\cref{thm:ilelinear} also yields methodological insight into the Structured prediction problem. Indeed, it materializes the gain obtained using ``structured'' losses such as the Hamming loss, over the $0-1$ loss, theoretically validating practitioner intuition. Furthermore, we show how prior information such as the bound on the input kernel $\kappa$ can have an impact of the chosen learning method.

\cref{thm:ilelinear} is of particular interest because the upper-bound is differentiable. In \cref{sec:algorithms}, we derive two learning algorithms that minimize the bound. The experimental results of this approach is presented in \cref{sec:experiments}.

\section{Experimental results}\label{sec:experiments}
In this paper, we prove a PAC-Bayes bound for Structured prediction using ILE losses. The bound provides an excess risk certificate that holds with high probability. Furthermore, we derive two learning algorithms from the bound, which both stem from minimizing a surrogate of the bound.

In this section, we present experimental results of three algorithms, on multi-label binary classification datasets.
All experiments can be reproduced with source code available\footnote{\github}.

\subsection{Algorithms}
First, we recall the different implement algorithms and their variants. We compare three families of algorithms:
\begin{itemize}
    \item \texttt{ILE($\lambda$)}: minimize \cref{eq:R-n-g} by Kernel Ridge Regression with regularization parameter $\lambda$. 
    \item \texttt{Relaxed-PB($\alpha, t$)}: minimize \cref{eq:ilelinobj} with gradient descent (with constant learning rate $\gamma$). This is described in \Cref{alg:gd-relax}.
    \item \texttt{MC-PB($\alpha, t$)}: minimize \cref{eq:optlintrue} with score-function gradient descent (with constant learning rate $\gamma$). 
\end{itemize}

For \texttt{Relaxed-PB} and \texttt{MC-PB}, $\alpha$ and $t$ are parameters of the prior $\mathcal P$ so are, in general, data-independent. See \cref{sec:methodology} for more details on the choice of $\alpha$ and $t$.

\subsection{Implementation}
In our work, we consider the PAC-Bayes framework: predictors $f$ are sampled from a posterior distribution $\mathcal Q$, itself learned from dataset $\mathcal S = \lbrace (x_1, y_1), \ldots, (x_m, y_m)\rbrace$. Recall that prior to learning, we define a prior distribution over predictors $\mathcal P$, which is independent from $\mathcal S$. The main difference between the PAC-Bayes approach and more general approaches is thus that we do not learn a predictor through empirical risk minimization but seek to estimate an optimal posterior distribution which minimizes a bound. In \cref{sec:algorithms} we showed how this can be reduced to minimizing a data-dependent penalized empirical risk objective.

Furthermore, in the ILE setting, learning a predictor $f: \mathcal X \rightarrow \mathcal Z$ (or predictor distribution $\mathcal Q$) is equivalent to learning a regressor $g:\mathcal X \rightarrow \mathcal H$ (or $\mathcal Q$). See \cref{fig:ile-schematic} for an overview. In \cref{sec:algorithms}, we present two alternative ways of learning $\mathcal Q$ over regressors $:g:\mathcal X \rightarrow \mathcal H$.

The combination of the ILE and PAC-Bayes settings in materialized in our implementation. Indeed, two main objects are implemented:

\begin{itemize}
    \item \texttt{Regressor} encodes data in $\mathcal X$ to $\mathcal H$, akin to $g$ (and to the mean of the posterior distribution $\mathcal Q$). The encoding is learned using one of the three families of algorithms mentioned above: \texttt{ILE}, \texttt{Relaxed-PB} and \texttt{MonteCarlo-PB}. As expected, in the case where \texttt{ILE} is employed, \texttt{Regressor} coincides with a Kernel ridge regressor.
    \item \texttt{StochasticPredictor} is a complete predictor from $\mathcal X$ to $\mathcal Z$, akin to $f\sim\mathcal Q$. It has a \texttt{Regressor} attribute and uses it to learn (\texttt{stochastic\_predictor.regressor.fit} is called) and perform inference. However, inference can be performed in one of two ways: deterministically from the mean of the posterior distribution $\mathcal Q$ (this reproduces the behavior of an ordinary predictor) or stochastically by sampling $M$ predictors $f_1, \ldots, f_m$ from $\mathcal Q$ and returning the $M$ corresponding predictions. Similarly, inference can be performed on regression by sampling $g_1, \ldots, g_m$ from $\mathcal Q$.
\end{itemize}

\begin{remark}
    Given $x\in\mathcal X$, our stochastic predictor returns $f_1(x) = d \circ g_1(x), \ldots, f_M(x) = d \circ g_M(x)$ where $g_1, \ldots, g_M$ are sampled iid from $\mathcal Q$. It is important to note that it does not return $\frac{1}{M}\sum_k f_k(x)$ (in the structured setting, this quantity is not defined) nor $d \circ \left(\frac{1}{M}\sum_k g_k(x)\right)$.
\end{remark}

\begin{remark}[Decoding]
    Recall that given an ILE loss $\Delta(z, y) = \langle \psi(y), \varphi(y) \rangle$ and a regressor $g$, the ILE framework defines $f$ by $f= d\circ g$ where:
    \begin{equation*}
        d(g(x)) = \arg\min_{z\in\mathcal Z} \langle\psi(z), g(x)\rangle.
    \end{equation*}
    We mentioned in \cref{sec:ile} that this decoder (or $\arg\min$ step) can be solved efficiently in certain cases. See for example \cite{nowak-vila-general,blondel_oracles,mensch_blondel}.
    In our implementation, $d$ is a naïve $\arg\max$ over all possible labels in $\mathcal Z$. Note however, that $d$ is not used during training.
\end{remark}

Both of these objects are implemented so as to be compatible with the popular \texttt{scikit-learn} library \citep{sklearn_api,scikit-learn}, in order to maximize their familiarity.

In our implementations of gradient descent and score-function gradient descent, we make use of automatic differentiation and safe pseudo-random number generation from the \texttt{jax} library \citep{jax2018github}. This presents several advantages: (i) our code is entirely written in \texttt{Python}, meaning it can be easily read and extended all the whilst being fast and memory efficient; (ii) reproduction is easy despite multiple, nested sampling steps that take place at training and at
inference; (iii) although not currently implemented, our code can be compiled to XLA and run on a GPU.

We use the \texttt{emotions} dataset from \texttt{scikit-multilearn}, a multi-label learning library for \texttt{Python} \citep{skml}. The dataset presents $m=593$ examples (we concatenate the training and test folds for our experiments), $\dim(\mathcal X) = 73$ and $\ell= 6$ binary labels. We use the linear kernel so $\mathcal F = \mathcal H$.

\subsection{Choice of the prior $\mathcal P(\alpha, t)$}\label{sec:methodology}
It is well-known that the choice of prior can help or hinder algorithm performance. In our case, the prior variance (we consider its mean to be $0$), which varies with $\alpha$ and $t$ determines the behaviors of the bound and derived algorithms.

\begin{figure}[t]
\begin{center}
    \includegraphics[width=0.8\textwidth]{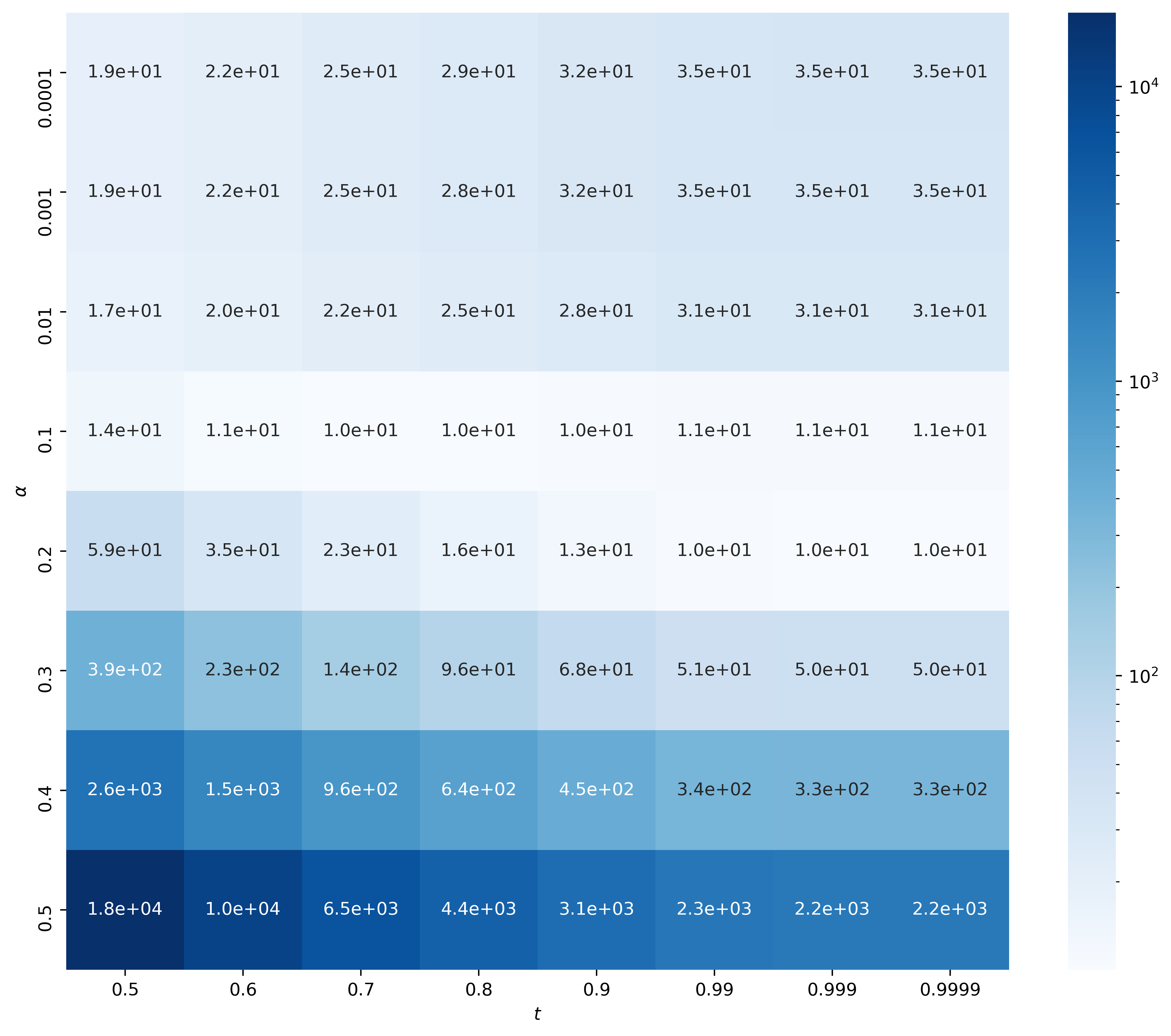}
\end{center}
\caption{$\hat J$ as a function of $\alpha$ and $t$ for the \texttt{ILE} algorithm. $\hat J$ is defined in \cref{eq:optlin}. The regularization parameter $\lambda$ is chosen to minimize $\hat J$.}
\label{fig:ile-loss-Q}
\end{figure}

\paragraph{Bound behavior} We examined the variation of the penalty term $\varepsilon^\prime$ in \cref{sec:consistency}. Here we consider the behavior of empirical risk term for the PAC-Bayes predictor associated to the ILE regressor. In this case, the mean of $\mathcal Q$ is independent of $\alpha$ and $t$. However, its variance $\sigma^2$ is chosen as a function of $\alpha$ and $t$. This impacts the bound value. Note that we cannot compute it explicitly (it depends on
intractable quantities such as $g^*$). In \cref{fig:ile-loss-Q}, we illustrate the dependence of the empirical risk part of the risk certificate $\hat J$, defined in \cref{eq:optlin}, on $\alpha$ and $t$. \cref{fig:ile-loss-Q} helps to illustrate the behavior of the ILE predictor in our PAC-Bayes setting, and gives an idea of how $\alpha$ and $t$ can impact the bound. Note that we are not offering this grid search experiment as a method of model selection. Indeed, the prior is assumed to be data independent and thus cannot be, without additional considerations, be optimized using bound values or predictor performance on the training set.

\paragraph{Algorithm behavior}
The choice of prior also impacts the behavior of the algorithms presented in this work: gradient descent on a relaxed objective or Monte Carlo gradient descent. Indeed, the relaxed objective presents terms which depend on the posterior variance. In the Monte Carlo approach, predictors are sampled from the posterior in order to optimize its mean. The behavior varies when $\sigma$ is large or small.

\begin{figure}[t]
\begin{center}
    \includegraphics[width=0.8\textwidth]{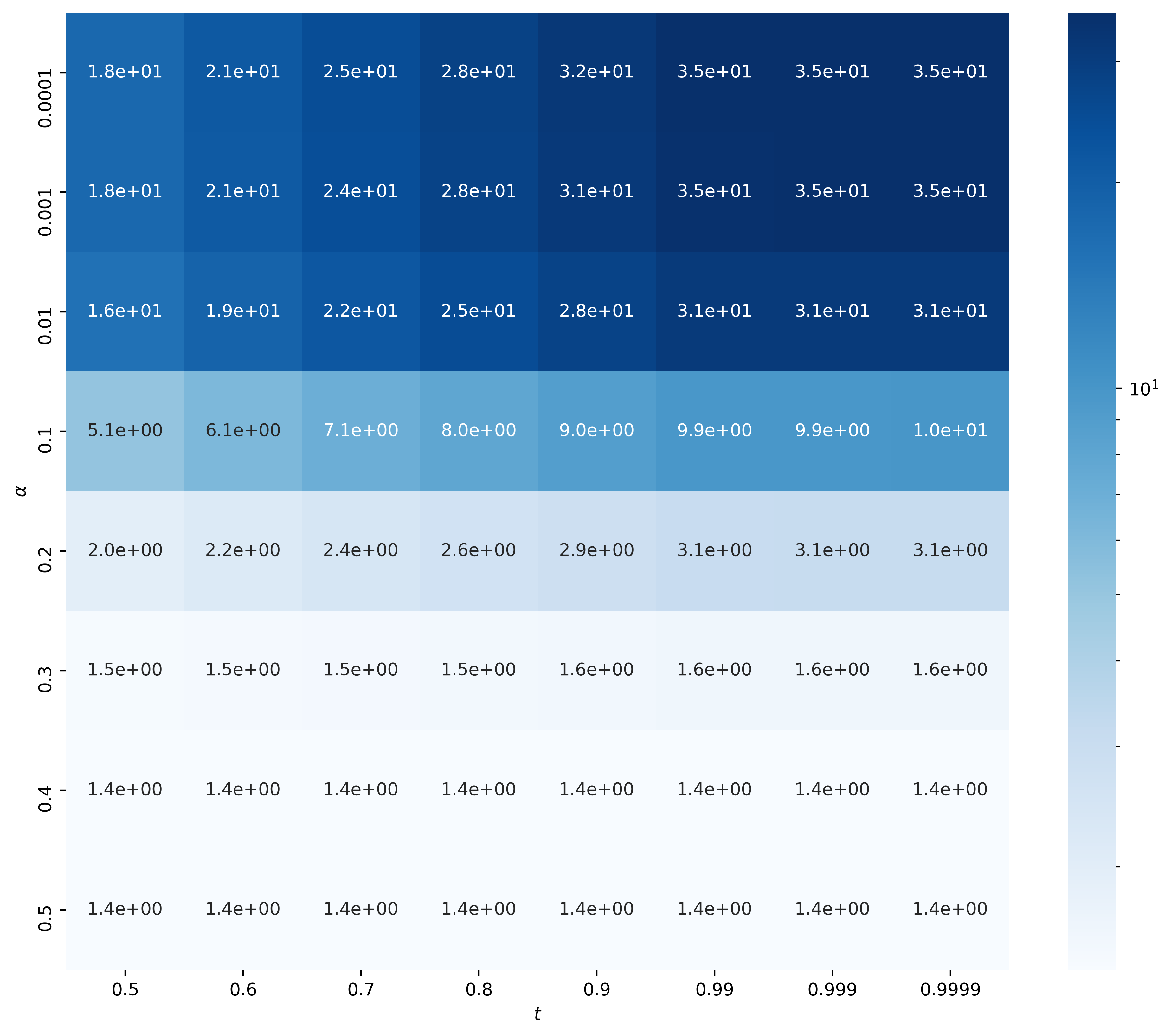}
\end{center}
\caption{$\hat J$ as a function of $\alpha$ and $t$ for the \texttt{Relaxed-PB} algorithm. $\hat J$ is defined in \cref{eq:optlin}. The learning rate is chosen between $\lbrace 10^{-8}, 10^{-4}, 10^{-3}, 10^{-2}\rbrace$ to minimize $\hat J$.}
\label{fig:relaxed-loss-Q}
\end{figure}

\begin{figure}[t]
\begin{center}
    \includegraphics[width=0.8\textwidth]{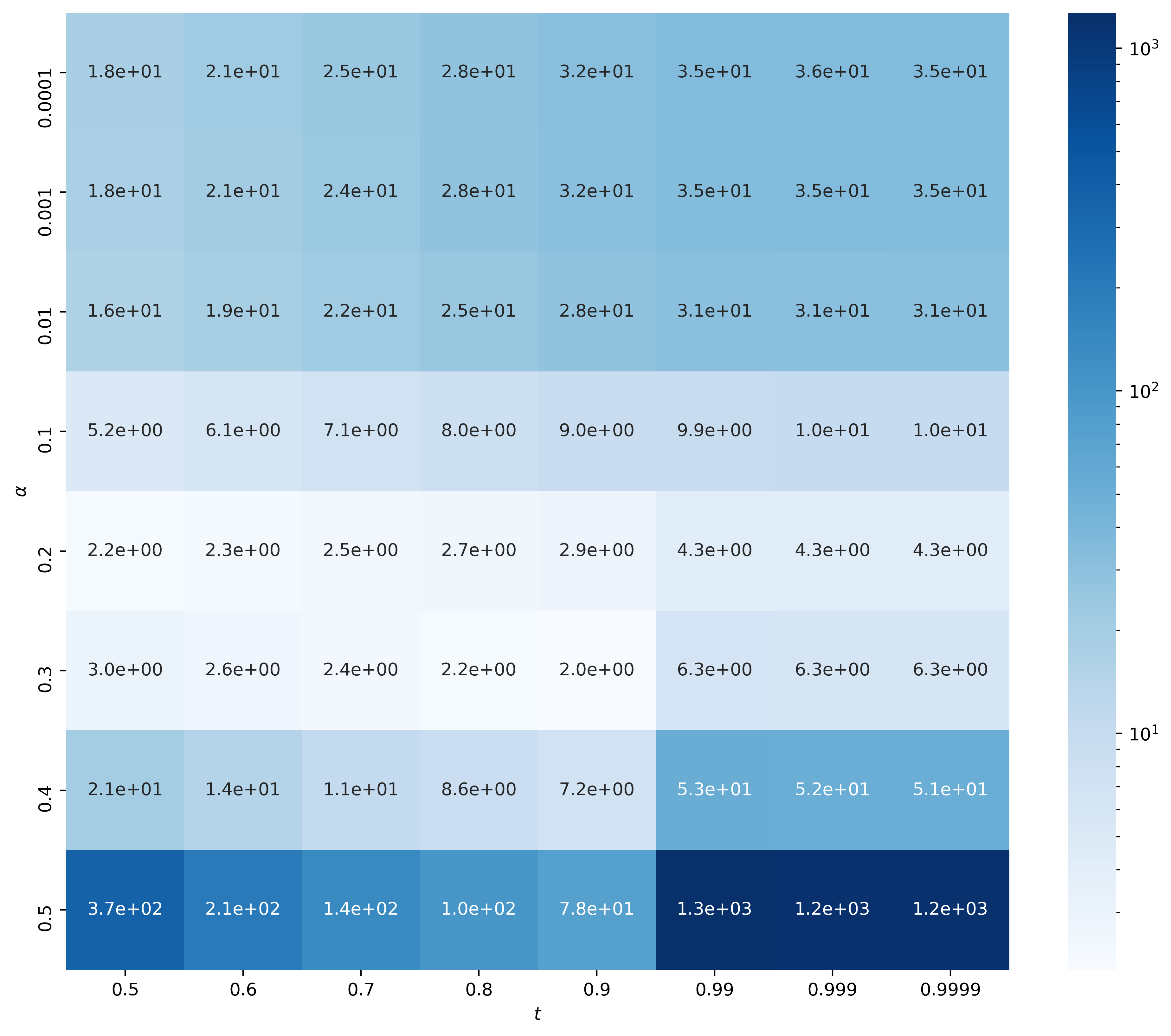}
\end{center}
\caption{$\hat J$ as a function of $\alpha$ and $t$ for the \texttt{MC-PB} algorithm. $\hat J$ is defined in \cref{eq:optlin}. The learning rate is chosen between $\lbrace 10^{-5}, 10^{-4}, 10^{-3}\rbrace$ to minimize $\hat J$. $M=20$ samples where used at each gradient step (with $\hat a =0$).}
\label{fig:mc-loss-Q}
\end{figure}

\begin{figure}[t]
    \centering
    \begin{minipage}{0.45\textwidth}
        \centering
        \includegraphics[width=\textwidth]{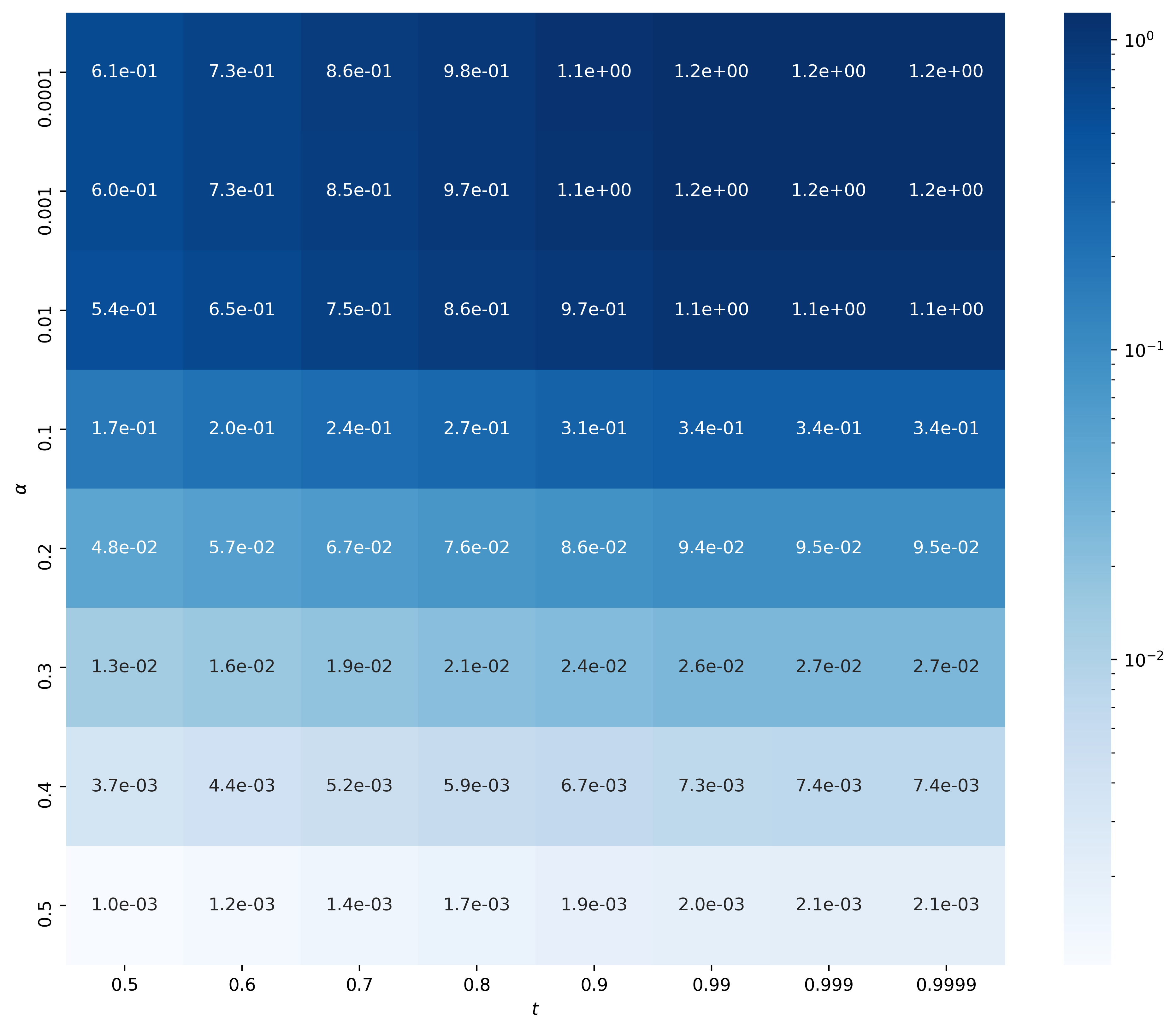}
        \caption{$\sigma$ as a function of $\alpha$ and $t$.}
        \label{fig:params-heatmap-1}
    \end{minipage}\hfill
    \begin{minipage}{0.45\textwidth}
        \centering
        \includegraphics[width=\textwidth]{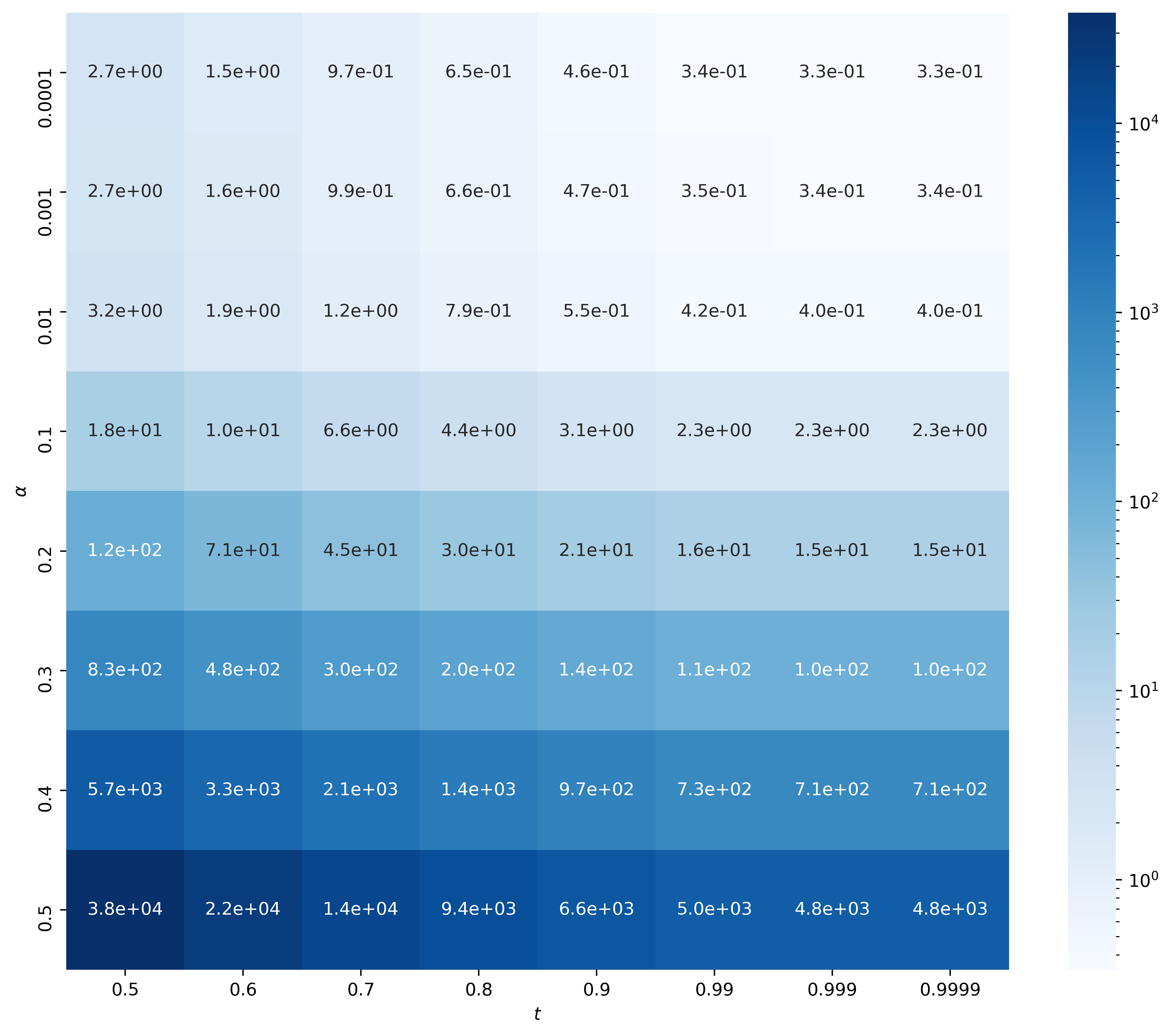}
        \caption{$\lambda$ as a function of $\alpha$ and $t$.}
        \label{fig:params-heatmap-2}
    \end{minipage}
\end{figure}

\cref{fig:relaxed-loss-Q,fig:mc-loss-Q} illustrates bound surrogate values ($\hat J$ is computed for all three algorithms) for different values of $t$ and $\alpha$. A complementary picture is given by the dependence of $\sigma$ and $\lambda_m^\alpha(t)$ on $\alpha$ and $t$, see \cref{fig:params-heatmap-1,fig:params-heatmap-2}.

\vskip 0.2in
\newpage
\bibliography{biblio}

\end{document}